\documentclass[letterpaper, 10 pt, conference]{ieeeconf}  
\usepackage{graphicx} 
\usepackage{amsmath}
\usepackage{bm}
\usepackage{amssymb}
\usepackage{booktabs}
\usepackage{tabularx}
\usepackage{multirow}
\usepackage{array}
\usepackage{stfloats}
\usepackage{ragged2e}
\usepackage{bbm}
\usepackage{balance}
\usepackage[colorlinks=true, linkcolor=blue, citecolor=blue, urlcolor=blue]{hyperref}
\usepackage[utf8]{inputenc}
\usepackage[style=ieee, backend=biber, sorting=none, minitems=2]{biblatex}
\IEEEoverridecommandlockouts                              
\title{\LARGE \bf
DKPMV: Dense Keypoints Fusion from Multi-View RGB Frames for 6D Pose Estimation of Textureless Objects
}

\author{%
Jiahong Chen$^{1, 2}$, Jinghao Wang$^{1, 2}$, Zi Wang$^{1, 2, *}$, Ziwen Wang$^{1, 2}$,  Banglei Guan$^{1, 2}$ and Qifeng Yu$^{1, 2}$%
\thanks{*Corresponding author: Zi Wang (wangzi16@nudt.edu.cn).}%
\thanks{$^{1}$College of Aerospace Science and Engineering, National University of Defense Technology, Changsha 410073, China.
$^{2}$Hunan Provincial Key Laboratory of Image Measurement and Vision Navigation, Changsha 410073, China.}%
}

\begin{document}

\maketitle
\thispagestyle{empty}
\pagestyle{empty}

\begin{abstract}

6D pose estimation of textureless objects is valuable for industrial robotic applications, yet remains challenging due to the frequent loss of depth information.
Current multi-view methods either rely on depth data or insufficiently exploit multi-view geometric cues, limiting their performance.
In this paper, we propose DKPMV, a pipeline that achieves dense keypoint-level fusion using only multi-view RGB images as input.
We design a three-stage progressive pose optimization strategy that leverages dense multi-view keypoint geometry information.
To enable effective dense keypoint fusion, we enhance the keypoint network with attentional aggregation and symmetry-aware training, improving prediction accuracy and resolving ambiguities on symmetric objects.
Extensive experiments on the ROBI dataset demonstrate that DKPMV outperforms state-of-the-art multi-view RGB approaches and even surpasses the RGB-D methods in the majority of cases.
The code will be available soon.
\end{abstract}

\section{INTRODUCTION}

Textureless objects are commonly encountered in modern industrial scenarios, such as mechanical components and plastic utensils~\cite{stoiber2022iterative}. 
The lack of distinctive color or texture makes these objects challenging for visual perception, drawing increasing attention in the robotics community~\cite{jin2023online},
as exemplified by benchmarks like the ROBI~\cite{yang2021robi} and XYZ-IBD datasets~\cite{huang2025xyz}. 
Accurate estimation of 6D poses is essential for a wide range of downstream robotic tasks, including autonomous grasping and assembly~\cite{shi2024asgrasp,zhang2024instance, kim2024sim,chen2025zerobp}.

In recent years, the vast majority of research has focused on addressing the 6D pose estimation problem of textureless objects using depth data~\cite{9197461,gao2021cloudaae,cai2022ove6d} or RGB-D images~\cite{wang2019densefusion,he2020pvn3ddeeppointwise3d,saadi2021optimizing}. 
While these approaches have significantly improved pose estimation accuracy, they heavily rely on the quality of the depth, which often degrades on specular surfaces~\cite{ni2025reasoninglearningperceptualmetric,yang2022next,junyang2023}, transparent materials~\cite{zhang2022transnet,liu2020keypose}, or low-light conditions~\cite{Chai2020DeepDepthFusion}.
Moreover, high-precision depth sensors are costly and operate at low frame rates, limiting their applicability in real-time robotic perception~\cite{cop2021new}.
RGB-based deep learning methods have effectively addressed 6D pose estimation for textureless objects~\cite{liu2024deeplearningbasedobjectpose}.
Nevertheless, the reliance on single-view input in most approaches~\cite{kleeberger2020single,su2022zebrapose,lian2023checkerpose} leads to limitations under scale ambiguity, occlusions, and symmetric object~\cite{bauer2024challenges}.

\begin{figure}[!t]
  \centering
  \refstepcounter{figure}
  \includegraphics[width=\linewidth]{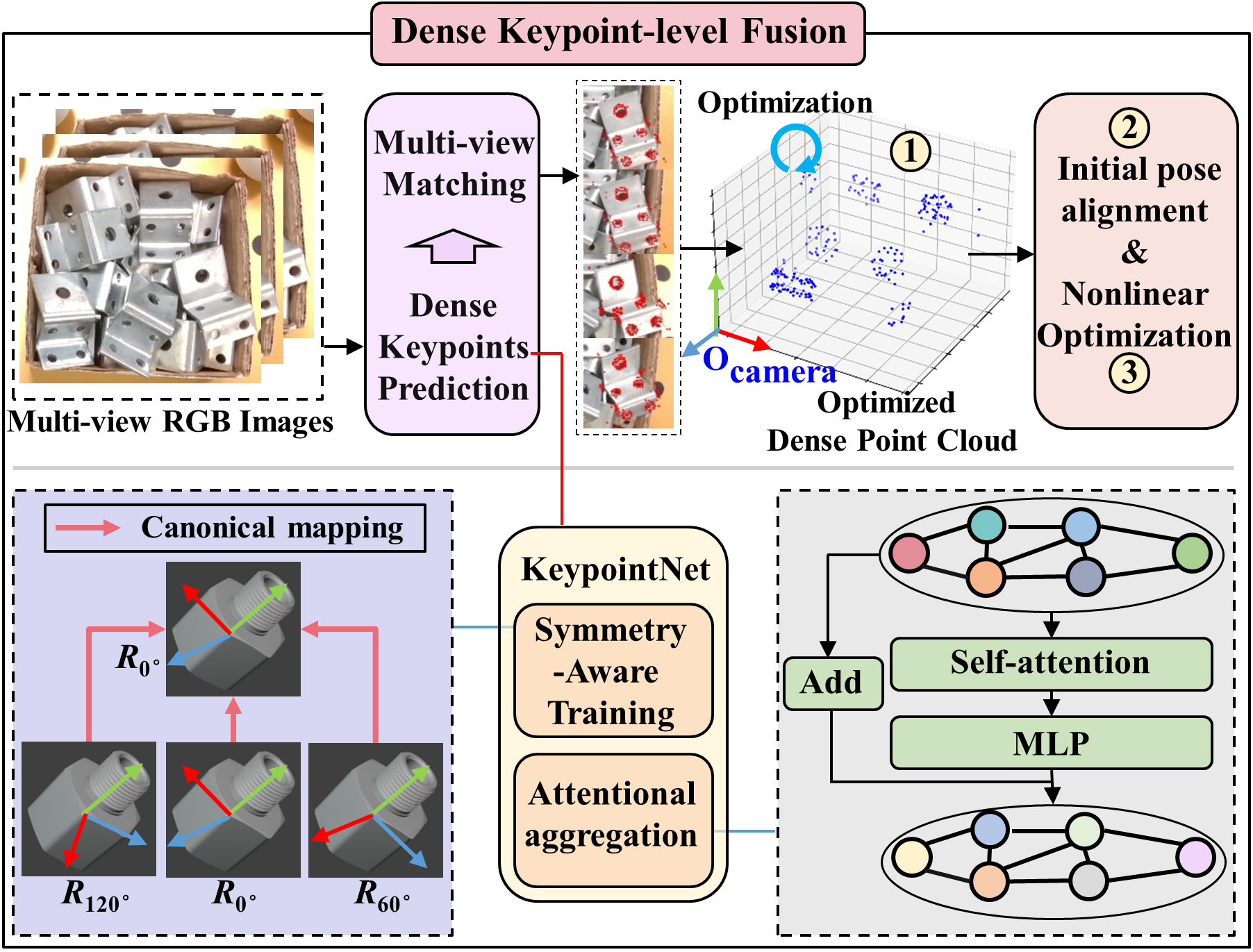}
  \label{fig:comparison}
  \vspace{-13pt}
  \begin{justify}
    \small Fig. 1: Brief overview of the DKPMV. A three-stage progressive optimization strategy is proposed, including dense point cloud reconstruction, initial pose alignment, and nonlinear optimization. To enable effective dense keypoint fusion, we enhance the keypoint network with attentional aggregation and symmetry-aware training.
  \end{justify}
\end{figure}

To address the inherent limitations of single RGB-view methods, recent researches have increasingly focused on multi-view frameworks for robust pose estimation of textureless objects. 
Existing approaches typically fall into two categories: sequential frame processing, as in object-level SLAM~\cite{fu2021multi,merrill2022symmetry,maninis2022vid2cad}, and simultaneous multi-frame input \cite{labbe2020cosypose,shugurov2021multi,LiMVKP, yang2025active}. 
While the former suffers from drift and latency due to frame-by-frame processing, the latter is more suitable for real-time robotic applications~\cite{choudhury2023tempo}. However, most existing multi-frame methods rely on single-view pose initialization followed by global refinement, resulting in pose-level fusion with limited exploitation of multi-view geometric consistency~\cite{labbe2020cosypose,shugurov2021multi,yang2025active}. 
While MV-3D-KP~\cite{LiMVKP} performs keypoint-level fusion to better utilize multi-view constraints, it integrates only sparse keypoints and remains dependent on depth input, which compromises robustness under occlusion or missing depth, as demonstrated in~\cite{yang2025active}.

Motivated by these challenges, we propose the DKPMV that performs dense keypoint-level multi-view fusion using only RGB images, as illustrated in Fig. \ref{fig:comparison}.
To resolve ambiguities in keypoint prediction for symmetric objects, we adopt a symmetry-aware training strategy (SAT)~\cite{pitteri2019object}.
Moreover, we also design an attentional aggregation module within the keypoint network to capture geometric constraints among dense keypoints and improve prediction accuracy.
Furthermore, we propose a three-stage progressive pose estimation strategy that leverages the dense keypoints.
These innovative designs enables reliable dense keypoint-level multi-view fusion and significantly improves the robustness of pose estimation in challenging scenarios. 
Extensive experiments on the challenging ROBI dataset~\cite{yang2021robi} demonstrate that DKPMV outperforms state-of-the-art RGB and RGB-D based approaches.

The main contributions of this work are as follows:
\begin{itemize}
    \item We propose the DKPMV, a 6D pose estimation framework that achieves dense keypoint-level multi-view fusion using only multiple RGB images as input.
    
    \item We employ SAT to resolve prediction ambiguities and design an attentional aggregation module to enhance keypoint accuracy and ensure effective fusion.
    
    \item We design a three-stage progressive pose optimization pipeline that integrates multi-view geometry with dense keypoint cues.
\end{itemize}

\section{RELATED WORK}

\subsection{6D Pose Estimation for Textureless Objects}
Traditional methods improve pose estimation accuracy for textureless objects by leveraging depth data~\cite{9197461,gao2021cloudaae,cai2022ove6d} or RGB-D images~\cite{wang2019densefusion,he2020pvn3ddeeppointwise3d,saadi2021optimizing}. 
Due to depth unreliability on textureless surfaces,
RGB-based approaches provide an efficient solution for 6D pose estimation by exploiting visual appearance cues~\cite{liu2024deeplearningbasedobjectpose}.
Template matching methods~\cite{deng2021poserbpf,sundermeyer2018implicit,li2020pose} infer poses by comparing images with pre-rendered templates and are suitable for textureless objects. 
Regression-based~\cite{hu2020single,wang2021gdr} methods estimate poses directly from global image features, offering efficiency. 
However, both approaches exhibit deteriorated accuracy when facing large viewpoint variations, occlusions, or domain shifts, due to the lack of explicit geometrical modeling.
Coordinates-based methods~\cite{xu20246d,liu2021kdfnet} aggregate pixel-wise object coordinates via RANSAC~\cite{fischler1981random}, ensuring robustness in occlusion.
Nonetheless, per-pixel coordinate regression takes vast computational burden, preventing their applications in multi-view and real-time scenarios. 

Semantic keypoint-based methods~\cite{pavlakos20176,liu2021mfpn,xu20246d} detect a set of 2D semantic keypoints and establish 2D–3D correspondences with predefined points on CAD models, from which the 6D pose is estimated via Perspective-n-Point (PnP).
This formulation transforms pose estimation into a structured and interpretable task, enabling consistent exploitation of multi-view geometric constraints~\cite{LiMVKP}.
Compared to sparse keypoints, dense keypoint-based methods~\cite{chen2022epro,di2021so,lian2023checkerpose,su2022zebrapose} capture richer scene information and exhibit greater robustness to occlusion.
Furthermore, integrating graph neural networks (GNN) enables geometric interactions among keypoints, further enhancing prediction accuracy~\cite{lian2023checkerpose}.
However, most existing methods utilize only single-view keypoint predictions without multi-view fusion, leading to depth loss and scale ambiguity. 
Moreover, they lack dedicated mechanisms to resolve keypoint ambiguities arising from object rotational symmetries~\cite{pitteri2019object}.

\subsection{Multi-View 6D Object Pose Estimation from RGB Images}
Multi-view 6D pose estimation has been shown to be effective in addressing the depth and scale ambiguities inherent in single-view settings. 
According to the input, these approaches can be categorized into sequential-frame based ones (e.g., object-level SLAM~\cite{fu2021multi,merrill2022symmetry,maninis2022vid2cad}) and simultaneous multi-frame based ones~\cite{labbe2020cosypose,shugurov2021multi,LiMVKP, yang2025active}. 
Sequential frame methods are susceptible to cumulative drift and processing latency, whereas simultaneous multi-frame method process multiple views in parallel, delivering the responsiveness required for real-time robotic tasks~\cite{choudhury2023tempo}.
However, most existing multi-frame methods perform pose-level fusion either through global optimization of individually estimated single-view poses~\cite{labbe2020cosypose,shugurov2021multi} or by decoupling translation and rotation estimation through multi-view center keypoint fusion~\cite{yang2025active}. 
Such strategies are prone to accumulated errors and limited in fully exploiting cross-view consistency. 
Keypoint-based methods offer a more interpretable and geometric formulation, enabling more geometry-aware multi-view integration~\cite{LiMVKP}. 
Nevertheless, current multi-view keypoint approaches are typically limited to sparse keypoints while requiring depth input, 
suffering from occlusion and degraded depth quality~\cite{LiMVKP}.

To address these challenges, we achieve dense keypoint fusion and introduce a three-stage progressive optimization pipeline that directly estimates 6D pose from RGB images, enabling robust and accurate keypoint-level multi-view fusion.
\begin{figure*}[t]
  \centering
  \refstepcounter{figure}
  \includegraphics[width=\linewidth]{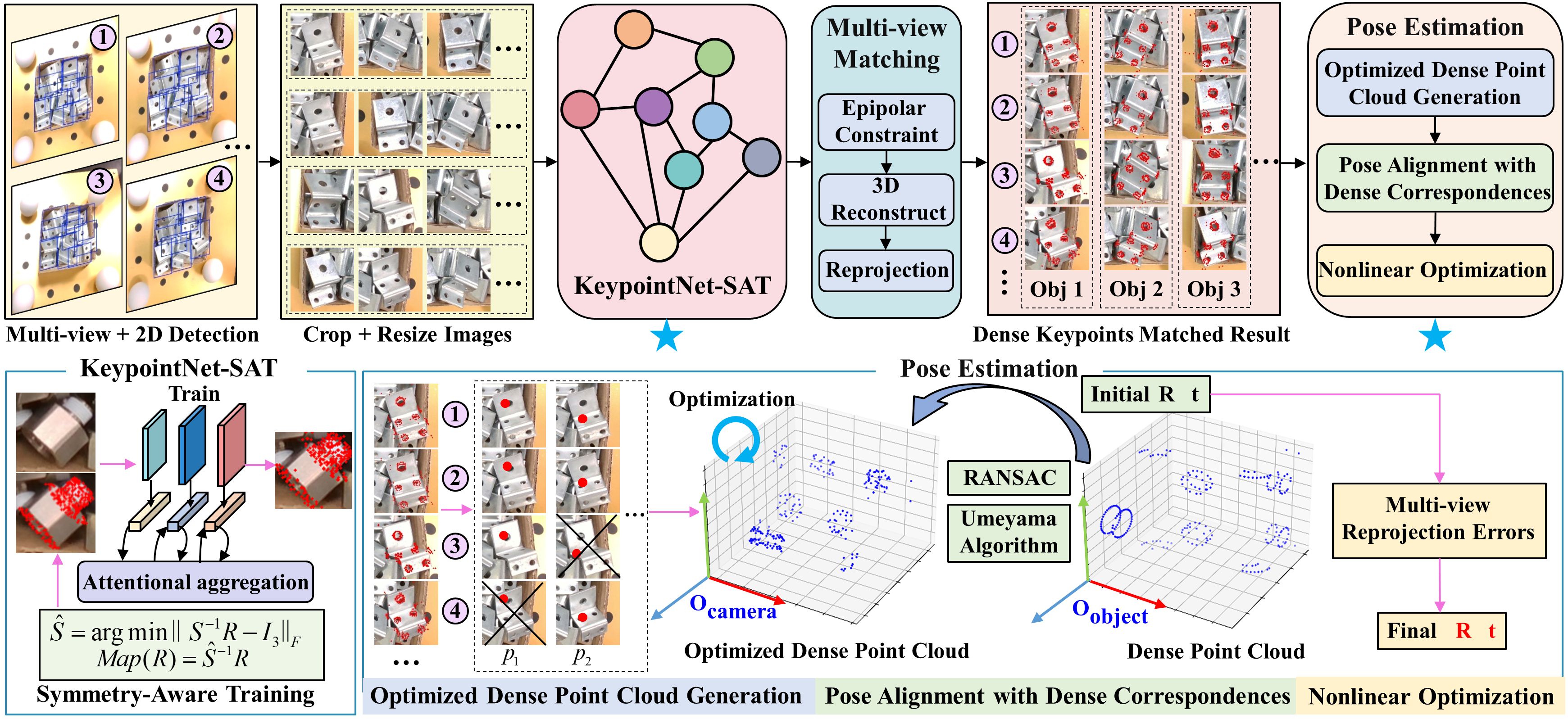}
  \label{fig:pipeline}
  \vspace{-13pt}
  \begin{justify}
    \small Fig. 2: Structural illustration of the DKPMV. Given multi-view RGB images and corresponding 2D bounding boxes, we perform dense keypoint detection for each object in each view. 
    The detected keypoints are matched across views and subsequently fed into a three-stage progressive pose estimation module.
  \end{justify}
\end{figure*}
\section{METHOD}
Given a set of RGB images $ \lbrace \mathcal I_{j} \rbrace_{j=1}^{V} $ and a collection of object instances, the objective is to estimate the 6D pose of each object, defined by its rotation $\mathbf{R} \in \mathrm{SO}(3)$ and translation $\mathbf{t} \in \mathbb{R}^3$, w.r.t. a global coordinate frame. We assume that the relative camera poses between views and the 3D geometry of each object (i.e., CAD models) are known. A set of $N$ 3D keypoints $ \boldsymbol{P}_{i}\in \mathbb{R}^{3} $ is uniformly sampled from each CAD model using farthest point sampling (FPS). 

As illustrated in Fig. \ref{fig:pipeline}, the proposed method consists of several stages, with the key innovation lying in the dense keypoint-level fusion across multiple views and a dedicated three-stage progressive pose optimization pipeline. 
We begin with 2D bounding box detection for each object instance in every image using an off-the-shelf YOLOv11. 
These bounding boxes are used to crop and resize the input images.
The patches are then processed by the KeypointNet-SAT network to generate dense keypoint predictions.
A multi-view matching module is then employed to establish consistent keypoint correspondences across views for each object instance. 
Finally, the matched dense keypoints are passed to the pose estimation module to recover the final 6D pose.

\subsection{Single-view Dense Keypoints Estimation}
The dense keypoint prediction network is built upon CheckerPose~\cite{lian2023checkerpose}.
It integrates a GNN to capture geometric relations among dense keypoints and a CNN to extract visual features from RGB inputs.
In addition to 2D keypoint coordinates, the network outputs a binary visibility code $ \bm{b}_{v} $ for each keypoint, indicating whether it lies within the object's region of interest (ROI). 
This visibility cue is used to filter out unreliable keypoints and improve the overall prediction quality.

To ensure effective keypoint fusion and stable network training, we adopt the SAT~\cite{pitteri2019object} to resolve keypoint prediction ambiguities on symmetric objects by transforming all equivalent poses into a unified canonical representation.
Specifically, given a proper symmetry group $ \mathcal{M}(O) $ for object $ O $, which defines the set of transformations that leave the object appearance unchanged:
\begin{equation}
\mathcal{M} = \{ \boldsymbol{m} \in \mathrm{SE}(3) \|
 \forall \boldsymbol{p} \in \mathrm{SE}(3), \mathcal{R}(O, \boldsymbol{p}) = \mathcal{R}(O, \boldsymbol{m} \cdot \boldsymbol{p}) \}
\end{equation}
where $ \mathcal{R}(O, \boldsymbol{p}) $ is the image of Object $ O $ under pose $ \boldsymbol{p} $  (ignoring lighting effects), $ \boldsymbol{m} $ is a rigid motion related to the symmetry, and $ \boldsymbol{m} \cdot \boldsymbol{p} $ is the pose after applying motion $ \boldsymbol{m} $. 
A corresponding operator $ \operatorname{Map}(\cdot) $ on $ SO(3) $ is employed to map symmetric equivalent poses to a consistent canonical form:
\begin{equation}
\operatorname{Map}(\mathbf{R})=\hat{S}^{-1} \mathbf{R}, \quad \forall \mathbf{R} \in \mathrm{SO}(3) \label{eq:4}
\end{equation}
where $ \operatorname{Map}(\cdot) $ ensures consistent keypoint supervision across all symmetric-equivalent poses during training. 
And $\hat{S}$ is the optimal rotation matrix that best aligns the input $\mathbf{R}$:
\begin{equation}
\hat{S}=\underset{S \in \mathcal{M}(O)}{\arg \min }\left\|S^{-1} \mathbf{R}-I_{3}\right\|_{F}
\end{equation}
By this means, we significantly improve pose estimation performance for rotationally symmetric objects, as discussed in Section~\ref{sec:experiments}.

Furthermore, inspired by the SuperGlue~\cite{sarlin2020superglue}, we replace the max-based node update scheme used in the EdgeConv~\cite{wang2019dynamic} module of CheckerPose with an attentional aggregation (Att), enabling more effective modeling of geometric relationships among dense keypoints.
Specifically, each keypoint $ \bm{x}_{i} $ aggregates features from its $ k=20 $ nearest neighbors $ \lbrace \bm{{x}}_{j}  \rbrace_{j\in \mathcal{N}(i)} $, as defined in CheckerPose~\cite{lian2023checkerpose}. The message $ \bm{m}_{\mathcal{E}\rightarrow i} $ is computed by attentional aggregation over all connected keypoints $ \lbrace j:(i,j)\in \mathcal{E} \rbrace $.
Given a query $ \bm{q}_{i} $ derived from the keypoint feature $\bm{x}_{i}^{\ell}$ at the $\ell$-th layer, and key and value $ \bm{k}_{j} $, $ \bm{v}_{j} $ derived from each neighboring keypoint feature $ \bm{x}_{j}^{\ell} $, the message is computed as a weighted average of the values:
\begin{equation}
\mathbf{m}_{\mathcal{E} \rightarrow i}=\sum_{j:(i, j) \in \mathcal{E}} \alpha_{i j} \mathbf{v}_{j}, \quad \alpha_{i j} = \text{Softmax}_{j} ( \mathbf{q}_{i}^{\top} \mathbf{k}_{j})
\end{equation}
and the updated keypoint feature at the $\ell+1$-th layer, denoted as $ \bm{x}_{i}^{\ell+1} $, is computed as:
\begin{equation}
\bm{x}_{i}^{\ell+1} = \bm{x}_{i}^{\ell} + \text{MLP} ( \mathbf{m}_{\mathcal{E} \rightarrow i})
\end{equation}
This node update strategy enhances keypoint localization accuracy, thereby improving the final pose estimation performance, as discussed in Section~\ref{sec:experiments}.

\subsection{Estimating Object Pose Using Matched Keypoints}
We accomplish multi-view keypoint matching by first pairing the two views with the closest poses, followed by keypoint reconstruction and projection of the reconstructed 3D points onto the remaining views. Implementation details of keypoint matching can be found in Appendix~\ref{app:1}. 

After matching instance-level keypoints across multiple views, we obtain a set of 2D predictions $ \bm{\tilde{p}}_{i}^{j} \in \mathbb{R}^2 $ for each instance, where $i$ indexes the keypoint and $j$ denotes the view. The theoretical projection of each keypoint $ \bm{p}_{i}^{j} \in \mathbb{R}^2 $ can be computed as:
\begin{equation}
\lambda_{i}^{j}\left[\begin{array}{c}\boldsymbol{p}_{i}^{j} \\ 1\end{array}\right]=\mathbf{K}_{j}\left(\mathbf{R}_{j}\left(\mathbf{R} \mathbf{P}_{i}+\mathbf{t}\right)+\mathbf{t}_{j}\right) \label{eq:6}
\end{equation}
where $\lambda_{i}^{j}$ denotes the depth value determined by the known camera intrinsics $\bm{K}_{j}$ and extrinsics $(\bm{R}_{j}, \bm{t}_{j})$, defined relative to the first camera.
Based on the Gaussian assumption and Bayes’ theorem, the optimal pose estimation can be formulated as:
\begin{equation}
\mathbf{R}^{*}, \mathbf{t}^{*}=\underset{\mathbf{R}, \boldsymbol{t}}{\arg \min } \sum_{i=1}^{\tilde{N}} \sum_{j=1}^{V}\left\|\tilde{\boldsymbol{p}}_{i}^{j}-\boldsymbol{p}_{i}^{j}\right\|_{2} \label{eq:7}
\end{equation}
where $\tilde{N}$ denotes the number of valid points that are simultaneously visible across all views, i.e., those with $\bm{b}_{v} = 1$, and thus satisfies $\tilde{N} \leq N$. Detailed derivation for Eq.~\ref{eq:7} is refered to Appendix~\ref{app:2}.

Based on Eq.~\ref{eq:6}, $\tilde{N}$ intermediate variables $ \bm{\tilde{P}}_i $ are introduced, which represent the reconstructed 3D keypoints and satisfies:
\begin{equation}
\mathbf{\tilde{P}}_i = \mathbf{R}\mathbf{P}_i + \mathbf{t},\quad i\in\lbrace 1,\ldots,\tilde{N} \rbrace \label{eq:8}
\end{equation}
Then, the optimization in Eq.~\ref{eq:7} can be equivalently decomposed into two subproblems:
\begin{align}
\tilde{\boldsymbol{P}}_{i}^{*} &=\underset{\tilde{\boldsymbol{P}}_{i}}{\arg \min } \sum_{j=1}^{V}\left\|\tilde{\boldsymbol{p}}_{i}^{j}-\boldsymbol{p}_{i}^{j}\right\|_{2}, i=\lbrace 1,\ldots, \tilde{N} \rbrace \label{eq:9}  \\
\mathbf{R}^{*}, \mathbf{t}^{*} &=\arg \min _{\mathbf{R}, \mathbf{t}} \sum_{i=1}^{\tilde{N}}\left\|\mathbf{R} \mathbf{P}_{i}+\mathbf{t}-\mathbf{\tilde{P}}_{i}^{*}\right\|_{2} \label{eq:10}
\end{align}

Motivated by Eqs.~\ref{eq:7}, ~\ref{eq:9}, and~\ref{eq:10}, 
we design a three-stage progressive optimization strategy for pose estimation.

\textbf{Optimized dense point cloud generation} 
We perform multiple iterations of RANSAC~\cite{fischler1981random}, where a threshold $\tau_{1}$ is applied, and select the candidate set of $\bm{\tilde{p}}_{i}^{j}$ with the best combined score of inlier count and reprojection error.
For each 3D point $\bm{\tilde{P}}_{i}$, we perform multi-view reconstruction using the valid view set $\mathcal{V}_{i}$ via singular value decomposition (SVD), resulting in the globally optimal estimate $\bm{\tilde{P}}_{i}^{*}$, as defined in Eq.~\ref{eq:9}.


\textbf{Pose alignment with dense correspondence} We estimate the initial pose $(\bm{R}^{*}, \bm{t}^{*})$ by aligning the reconstructed dense point cloud $ \mathcal{P}=\lbrace \bm{\tilde{P}}_{i}^{*}\in \mathbb{R}^{3}\rbrace_{i=1}^{\tilde{N}} $ with the reference 3D keypoints $ \mathcal{P}_{o}=\lbrace \bm{P}_{i}\in \mathbb{R}^{3}\rbrace_{i=1}^{\tilde{N}} $.
Specifically, we solve Eq.~\ref{eq:10} using the Umeyama algorithm~\cite{umeyama2002least}, combined with RANSAC~\cite{fischler1981random}, where a threshold $\tau_{2}$ is imposed to suppress outliers by selecting the solution with the maximum number of inliers from $\mathcal{P}$.

\textbf{Nonlinear optimization} Given the initial pose estimates ($ \mathbf{R}^{*},\mathbf{t}^{*} $), we further refine the solution by performing global nonlinear optimization (NO), leveraging multi-view keypoint predictions.

To mitigate the influence of outliers, only valid 2D keypoints are included in the optimization, and the cost function is defined as:
\begin{equation}
\underset{\mathbf{R^{*}}, \mathbf{t^{*}}}{\arg \min } \sum_{i=1}^{N_{\text{inlier}}^{3d}} \sum_{j \in \mathcal{V}_i} \sigma\left(\left\|\mathbf{p}_{i}^{j}-\mathbf{\tilde{p}}_{i}^{j}\right\|_{2}\right)
\end{equation}
where $ \sigma(\cdot) $ denotes a robust loss function to reduce the impact of heavy-tailed noise.

Please refer to Appendix~\ref{app:3} for more detailed information on the three-stage progressive pose estimation method.

\section{EXPERIMENTS}
\label{sec:experiments}

\subsection{Datasets and Evaluation Metrics}

We evaluate our method on the ROBI dataset~\cite{yang2021robi}, which provides multi-view RGB images and ground-truth 6D poses for 7 textureless industrial objects.
The dataset comprises images acquired using a high-end Ensenso and a low-cost RealSense sensor.
To balance speed and accuracy, we adopt 256 keypoints as our primary configuration.
Following~\cite{yang2025active,junyang2023}, we train our keypoint network with only synthetic images provided by the ROBI dataset.
Evaluation is conducted on the real-world test sets captured by both Ensenso and RealSense sensors for all  objects.

Following~\cite{yang2025active}, we evaluate pose estimation performance using the common average recall (AR) under two adopted metrics: the average distance (ADD) and the 5-mm/10-degree ($5\mathrm{mm}, 10^\circ$) metric.
For symmetric objects, we report the minimum error over all equivalent ground-truth poses under symmetry, based on either the ADD or the ($5\mathrm{mm}, 10^\circ$) metric.
In our evaluation, a ground-truth pose is considered valid only if its visibility score is larger than $ 75\% $, consistent with the rule in~\cite{yang2025active}.

\subsection{Comparison with State-of-the-Art Methods}
\begin{table*}[!th]
	\centering
    \refstepcounter{table}
    \label{tab:1}
    \renewcommand{\arraystretch}{1.03}
	\begin{tabularx}{\textwidth}{>{\centering\arraybackslash}p{32pt}>{\centering\arraybackslash}X>{\centering\arraybackslash}X>{\centering\arraybackslash}X>{\centering\arraybackslash}p{40pt}>{\centering\arraybackslash}X>{\centering\arraybackslash}X|>{\centering\arraybackslash}X>{\centering\arraybackslash}X>{\centering\arraybackslash}p{40pt}>{\centering\arraybackslash}X>{\centering\arraybackslash}X}
		\hline
		\multicolumn{2}{c}{\multirow{2}{*}{Objects}}                & \multicolumn{5}{c|}{4 Views}                                                                                                                   & \multicolumn{5}{c}{8 Views}                \\ \cline{3-12}
		\multicolumn{2}{c}{}                                        & \multicolumn{2}{c}{CosyPose + LINE2D}                 & MV-3D-KP                  & Jun’s                     & Ours                       & \multicolumn{2}{c}{CosyPose + LINE2D}     & MV-3D-KP & Jun’s  & Ours   \\ \hline
		\multicolumn{2}{c}{Input Modality}                          & RGB                       & RGBD                      & RGBD                      & RGB                       & RGB                        & RGB                 & RGBD                & RGBD     & RGB    & RGB    \\ \hline
		Tube$^{*}$                      & \multicolumn{1}{|c}{ADD}     & \multicolumn{1}{|c}{32.5} & \multicolumn{1}{|c}{74.8} & \multicolumn{1}{|c}{\textbf{94.0}} & \multicolumn{1}{|c}{89.4} & \multicolumn{1}{|c|}{92.1} & 50.3                & 91.4                & \textbf{96.0}     & 94.0   & 94.7   \\
		Fitting$^{\dag}$                  & \multicolumn{1}{|c}{\text{(5, 10)}} & \multicolumn{1}{|c}{45.7} & \multicolumn{1}{|c}{76.2} & \multicolumn{1}{|c}{\textbf{95.4}} & \multicolumn{1}{|c}{88.1} & \multicolumn{1}{|c|}{89.4} & 71.5                & 94.7                & \textbf{96.0}     & 92.0   & \textbf{96.0}   \\ \hline
		Chrome$^{*}$                    & \multicolumn{1}{|c}{ADD}     & \multicolumn{1}{|c}{55.7} & \multicolumn{1}{|c}{73.0} & \multicolumn{1}{|c}{90.8} & \multicolumn{1}{|c}{86.7} & \multicolumn{1}{|c|}{\textbf{96.4}} & 70.1                & 88.5                & 91.9     & 93.7   & \textbf{98.2}   \\
		 Screw$^{\dag}$                    & \multicolumn{1}{|c}{\text{(5, 10)}} & \multicolumn{1}{|c}{63.2} & \multicolumn{1}{|c}{78.2} & \multicolumn{1}{|c}{88.5} & \multicolumn{1}{|c}{85.1} & \multicolumn{1}{|c|}{\textbf{92.9}} & 78.7                & 90.2                & 90.8     & 90.2   & \textbf{97.6}   \\ \hline
		\multirow{2}{*}{Eye Bolt$_{\dag}^{*}$} & \multicolumn{1}{|c}{ADD}     & \multicolumn{1}{|c}{35.1} & \multicolumn{1}{|c}{85.1} & \multicolumn{1}{|c}{\textbf{93.2}} & \multicolumn{1}{|c}{\textbf{93.2}} & \multicolumn{1}{|c|}{90.5} & 79.7                & 93.2                & \textbf{94.6}     & \textbf{94.6}   & 90.5   \\
		                          & \multicolumn{1}{|c}{\text{(5, 10)}} & \multicolumn{1}{|c}{27.0} & \multicolumn{1}{|c}{78.4} & \multicolumn{1}{|c}{\textbf{87.8}} & \multicolumn{1}{|c}{67.6} & \multicolumn{1}{|c|}{81.1} & 64.9                & 83.8                & \textbf{85.1}     & 75.8   & \textbf{85.1}   \\ \hline
		\multirow{2}{*}{Gear$^{*}$}     & \multicolumn{1}{|c}{ADD}     & \multicolumn{1}{|c}{25.9} & \multicolumn{1}{|c}{80.2} & \multicolumn{1}{|c}{85.2} & \multicolumn{1}{|c}{91.4} & \multicolumn{1}{|c|}{\textbf{98.8}} & 43.2                & 88.9                & 93.8     & 97.5   & \textbf{100.0}  \\
		                          & \multicolumn{1}{|c}{\text{(5, 10)}} & \multicolumn{1}{|c}{29.6} & \multicolumn{1}{|c}{79.0} & \multicolumn{1}{|c}{85.2} & \multicolumn{1}{|c}{85.2} & \multicolumn{1}{|c|}{\textbf{92.6}} & 45.7                & 92.6                & 91.4     & 93.8   & \textbf{96.3}   \\ \hline
		\multirow{2}{*}{Zigzag}   & \multicolumn{1}{|c}{ADD}     & \multicolumn{1}{|c}{65.5} & \multicolumn{1}{|c}{87.9} & \multicolumn{1}{|c}{96.6} & \multicolumn{1}{|c}{94.8} & \multicolumn{1}{|c|}{\textbf{100.0}} & 77.6                & 96.6                & 96.6     & 98.3   & \textbf{100.0}   \\
		                          & \multicolumn{1}{|c}{\text{(5, 10)}} & \multicolumn{1}{|c}{37.9} & \multicolumn{1}{|c}{75.9} & \multicolumn{1}{|c}{93.1} & \multicolumn{1}{|c}{89.7} & \multicolumn{1}{|c|}{\textbf{100.0}} & 63.8                & 93.1                & 96.6     & 93.1   & \textbf{100.0}   \\ \hline
		DIN                       & \multicolumn{1}{|c}{ADD}     & \multicolumn{1}{|c}{15.6}  & \multicolumn{1}{|c}{57.8} & \multicolumn{1}{|c}{90.6} & \multicolumn{1}{|c}{69.5} & \multicolumn{1}{|c|}{\textbf{92.1}} & 24.2                & 64.1                & \textbf{93.8}     & 73.4   & 92.9   \\
		Connector                 & \multicolumn{1}{|c}{\text{(5, 10)}} & \multicolumn{1}{|c}{12.5}  & \multicolumn{1}{|c}{46.1} & \multicolumn{1}{|c}{84.4} & \multicolumn{1}{|c}{53.9} & \multicolumn{1}{|c|}{\textbf{89.7}} & 23.4                 & 51.6                & \textbf{93.0}     & 59.4   & 92.9   \\ \hline
		D-Sub$^{*}$                     & \multicolumn{1}{|c}{ADD}     & \multicolumn{1}{|c}{9.9}  & \multicolumn{1}{|c}{55.3} & \multicolumn{1}{|c}{\textbf{92.5}} & \multicolumn{1}{|c}{79.5} & \multicolumn{1}{|c|}{81.4} & 15.5                 & 63.3                & \textbf{95.7}     & 84.5   & 81.4   \\
		Connector$^{\dag}$                 & \multicolumn{1}{|c}{\text{(5, 10)}} & \multicolumn{1}{|c}{11.2}  & \multicolumn{1}{|c}{39.1} & \multicolumn{1}{|c}{\textbf{83.2}} & \multicolumn{1}{|c}{47.2} & \multicolumn{1}{|c|}{75.2} & 11.2                 & 41.6                & \textbf{91.3}     & 55.9   & 76.4   \\ \hline
		\multirow{2}{*}{ALL}      & \multicolumn{1}{|c}{ADD}     & \multicolumn{1}{|c}{34.3} & \multicolumn{1}{|c}{73.4} & \multicolumn{1}{|c}{91.8} & \multicolumn{1}{|c}{86.4} & \multicolumn{1}{|c|}{\textbf{93.0}} & 51.5                & 83.7                & \textbf{94.6}     & 90.9   & 94.0   \\
		                          & \multicolumn{1}{|c}{\text{(5, 10)}} & \multicolumn{1}{|c}{32.4} & \multicolumn{1}{|c}{67.6} & \multicolumn{1}{|c}{88.2} & \multicolumn{1}{|c}{73.8} & \multicolumn{1}{|c|}{\textbf{88.7}} & 51.3                & 78.2                & \textbf{92.0}     & 80.0   & \textbf{92.0}   \\ \hline                                             
	\end{tabularx}
    \vspace{-1pt}
    \begin{justify}
        \small Table \uppercase\expandafter{\romannumeral1}: The AR (\%) of 6D object pose estimation on \textbf{Ensenso} test set, evaluated with the metrics of ADD and ($5\mathrm{mm}, 10^\circ$). There are a total of nine scenes for each object. (${\dag}$) indicates the use of the SAT strategy, and (${*}$) denotes symmetric objects, following~\cite{yang2025active}.
    \end{justify}
\end{table*}
\begin{table*}[!th]
	\centering
    \refstepcounter{table}
    \label{tab:2}
    \renewcommand{\arraystretch}{1.03}
	\begin{tabularx}{\textwidth}{>{\centering\arraybackslash}p{32pt}>{\centering\arraybackslash}X>{\centering\arraybackslash}X>{\centering\arraybackslash}X>{\centering\arraybackslash}p{40pt}>{\centering\arraybackslash}X>{\centering\arraybackslash}X|>{\centering\arraybackslash}X>{\centering\arraybackslash}X>{\centering\arraybackslash}p{40pt}>{\centering\arraybackslash}X>{\centering\arraybackslash}X}
		\hline
		\multicolumn{2}{c}{\multirow{2}{*}{Objects}}                & \multicolumn{5}{c|}{4 Views}                                                                                                                   & \multicolumn{5}{c}{8 Views}                \\ \cline{3-12}
		\multicolumn{2}{c}{}                                        & \multicolumn{2}{c}{CosyPose + LINE2D}                 & MV-3D-KP                  & Jun’s                     & Ours                       & \multicolumn{2}{c}{CosyPose + LINE2D}     & MV-3D-KP & Jun’s  & Ours   \\ \hline
		\multicolumn{2}{c}{Input Modality}                          & RGB                       & RGBD                      & RGBD                      & RGB                       & RGB                        & RGB                 & RGBD                & RGBD     & RGB    & RGB    \\ \hline
		Tube$^{*}$                      & \multicolumn{1}{|c}{ADD}     & \multicolumn{1}{|c}{27.9} & \multicolumn{1}{|c}{70.6} & \multicolumn{1}{|c}{80.9} & \multicolumn{1}{|c}{\textbf{86.8}} & \multicolumn{1}{|c|}{83.8} & 69.1                & 83.9                & 82.4     & 85.3   & \textbf{95.6}   \\
		Fitting$^{\dag}$                   & \multicolumn{1}{|c}{\text{(5, 10)}} & \multicolumn{1}{|c}{48.5} & \multicolumn{1}{|c}{72.1} & \multicolumn{1}{|c}{67.6} & \multicolumn{1}{|c}{79.4} & \multicolumn{1}{|c|}{\textbf{82.4}} & 82.3                & 85.3                & 70.6     & 91.2   & \textbf{95.6}   \\ \hline
		Chrome$^{*}$                    & \multicolumn{1}{|c}{ADD}     & \multicolumn{1}{|c}{58.6} & \multicolumn{1}{|c}{68.5} & \multicolumn{1}{|c}{78.6} & \multicolumn{1}{|c}{92.9} & \multicolumn{1}{|c|}{\textbf{95.7}} & 77.1                & 80.0                & 84.3     & 92.9   & \textbf{95.7}   \\
		 Screw$^{\dag}$                    & \multicolumn{1}{|c}{\text{(5, 10)}} & \multicolumn{1}{|c}{64.3} & \multicolumn{1}{|c}{82.9} & \multicolumn{1}{|c}{80.0} & \multicolumn{1}{|c}{77.1} & \multicolumn{1}{|c|}{\textbf{94.3}} & 85.7                & 94.3                & 90.0     & 87.1   & \textbf{95.7}   \\ \hline
		\multirow{2}{*}{Eye Bolt$_{\dag}^{*}$} & \multicolumn{1}{|c}{ADD}     & \multicolumn{1}{|c}{58.8} & \multicolumn{1}{|c}{76.5} & \multicolumn{1}{|c}{88.2} & \multicolumn{1}{|c}{\textbf{94.1}} & \multicolumn{1}{|c|}{91.2} & 73.5                & \textbf{94.1}                & 85.3     & \textbf{94.1}   & 91.2   \\
		                          & \multicolumn{1}{|c}{\text{(5, 10)}} & \multicolumn{1}{|c}{41.2} & \multicolumn{1}{|c}{67.6} & \multicolumn{1}{|c}{\textbf{79.4}} & \multicolumn{1}{|c}{55.9} & \multicolumn{1}{|c|}{73.5} & 61.8                & \textbf{91.2}                & 79.4     & 76.5   & 85.3   \\ \hline
		\multirow{2}{*}{Gear$^{*}$}     & \multicolumn{1}{|c}{ADD}     & \multicolumn{1}{|c}{36.1} & \multicolumn{1}{|c}{83.3} & \multicolumn{1}{|c}{80.6} & \multicolumn{1}{|c}{94.4} & \multicolumn{1}{|c|}{\textbf{97.2}} & 55.6                & 97.2                & 88.9     & 97.2   & \textbf{100.0}  \\
		                          & \multicolumn{1}{|c}{\text{(5, 10)}} & \multicolumn{1}{|c}{38.9} & \multicolumn{1}{|c}{77.8} & \multicolumn{1}{|c}{52.8} & \multicolumn{1}{|c}{86.1} & \multicolumn{1}{|c|}{\textbf{97.2}} & 58.3                & 94.4                & 72.2     & 88.9   & \textbf{97.2}   \\ \hline
		\multirow{2}{*}{Zigzag}   & \multicolumn{1}{|c}{ADD}     & \multicolumn{1}{|c}{42.9} & \multicolumn{1}{|c}{78.6} & \multicolumn{1}{|c}{\textbf{96.4}} & \multicolumn{1}{|c}{89.3} & \multicolumn{1}{|c|}{92.9} & 71.4                & 92.9                & \textbf{96.4}     & \textbf{96.4}   & \textbf{96.4}   \\
		                          & \multicolumn{1}{|c}{\text{(5, 10)}} & \multicolumn{1}{|c}{21.4} & \multicolumn{1}{|c}{71.4} & \multicolumn{1}{|c}{\textbf{92.9}} & \multicolumn{1}{|c}{85.7} & \multicolumn{1}{|c|}{\textbf{92.9}} & 64.3                & 92.9                & \textbf{96.4}     & 92.9   & \textbf{96.4}   \\ \hline
		DIN                       & \multicolumn{1}{|c}{ADD}     & \multicolumn{1}{|c}{3.8}  & \multicolumn{1}{|c}{36.5} & \multicolumn{1}{|c}{86.5} & \multicolumn{1}{|c}{51.9} & \multicolumn{1}{|c|}{\textbf{98.1}} & 15.1                & 51.9                & 84.6     & 82.7   & \textbf{98.1}   \\
		Connector                 & \multicolumn{1}{|c}{\text{(5, 10)}} & \multicolumn{1}{|c}{1.9}  & \multicolumn{1}{|c}{30.8} & \multicolumn{1}{|c}{76.9} & \multicolumn{1}{|c}{32.7} & \multicolumn{1}{|c|}{\textbf{86.5}} & 9.6                 & 34.6                & 84.6     & 57.7   & \textbf{96.2}   \\ \hline
		D-Sub$^{*}$                     & \multicolumn{1}{|c}{ADD}     & \multicolumn{1}{|c}{6.9}  & \multicolumn{1}{|c}{40.3} & \multicolumn{1}{|c}{81.9} & \multicolumn{1}{|c}{70.8} & \multicolumn{1}{|c|}{\textbf{88.9}} & 9.7                 & 45.8                & 83.3     & 81.9   & \textbf{91.7}   \\
		Connector$^{\dag}$                 & \multicolumn{1}{|c}{\text{(5, 10)}} & \multicolumn{1}{|c}{6.9}  & \multicolumn{1}{|c}{18.1} & \multicolumn{1}{|c}{45.8} & \multicolumn{1}{|c}{31.9} & \multicolumn{1}{|c|}{\textbf{83.3}} & 8.3                 & 33.3                & 43.1     & 43.1   & \textbf{88.9}   \\ \hline
		\multirow{2}{*}{ALL}      & \multicolumn{1}{|c}{ADD}     & \multicolumn{1}{|c}{33.6} & \multicolumn{1}{|c}{64.9} & \multicolumn{1}{|c}{84.7} & \multicolumn{1}{|c}{82.9} & \multicolumn{1}{|c|}{\textbf{92.5}} & 53.1                & 78.0                & 86.5     & 90.1   & \textbf{95.5}   \\
		                          & \multicolumn{1}{|c}{\text{(5, 10)}} & \multicolumn{1}{|c}{31.9} & \multicolumn{1}{|c}{60.1} & \multicolumn{1}{|c}{70.8} & \multicolumn{1}{|c}{64.1} & \multicolumn{1}{|c|}{\textbf{87.2}} & 52.9                & 75.1                & 76.6     & 76.8   & \textbf{93.6}   \\ \hline                     
	\end{tabularx}
    \vspace{-1pt}
    \begin{justify}
        \small Table \uppercase\expandafter{\romannumeral2}: The AR (\%) of 6D object pose estimation on \textbf{RealSense} test set, evaluated with the metrics of ADD and ($5\mathrm{mm}, 10^\circ$). There are a total of four scenes for each object. (${\dag}$) indicates the use of the SAT strategy, and (${*}$) denotes symmetric objects, following~\cite{yang2025active}.
    \end{justify}
\end{table*}

Following~\cite{yang2025active}, we conduct experiments on the ROBI dataset using varying numbers of views (4 and 8).
We compare our method with SOTA multi-view approaches, including both RGB and RGB-D methods: CosyPose~\cite{labbe2020cosypose}, MV-3D-KP~\cite{LiMVKP}, and Jun's method~\cite{yang2025active}.
Tables~\ref{tab:1} and~\ref{tab:2} report the pose estimation results for seven highly reflective textureless objects on the Ensenso and RealSense test sets, respectively.

On the \textbf{Ensenso} test set, our method achieves the highest AR under the 4-view setting, reaching 93.0\% and 88.7\% on the ADD and $(5\mathrm{mm}, 10^\circ)$ metrics respectively, surpassing the RGB-D-based MV-3D-KP method.
With 8 input views, MV-3D-KP achieves the best overall performance, benefiting from high quality depth data.
Nevertheless, our method also improves over the 4-views setting, achieving competitive accuracy and falling only slightly behind MV-3D-KP's 94.6\% under the ADD metric, while still outperforming all RGB-based methods.

On the \textbf{RealSense} test set, the performance of RGB-D methods drops significantly due to the reduced quality of depth data, whereas our method demonstrates a substantial advantage.
Its performance significantly surpasses both RGB and RGB-D methods, especially under the $(5\mathrm{mm}, 10^\circ)$ metric, where it outperforms the second-best method by a large margin of 16.4\% and 16.8\% in the 4-view and 8-view settings respectively.

\begin{figure}[!ht]
	\centering
    \refstepcounter{figure}
	\includegraphics[width=\linewidth]{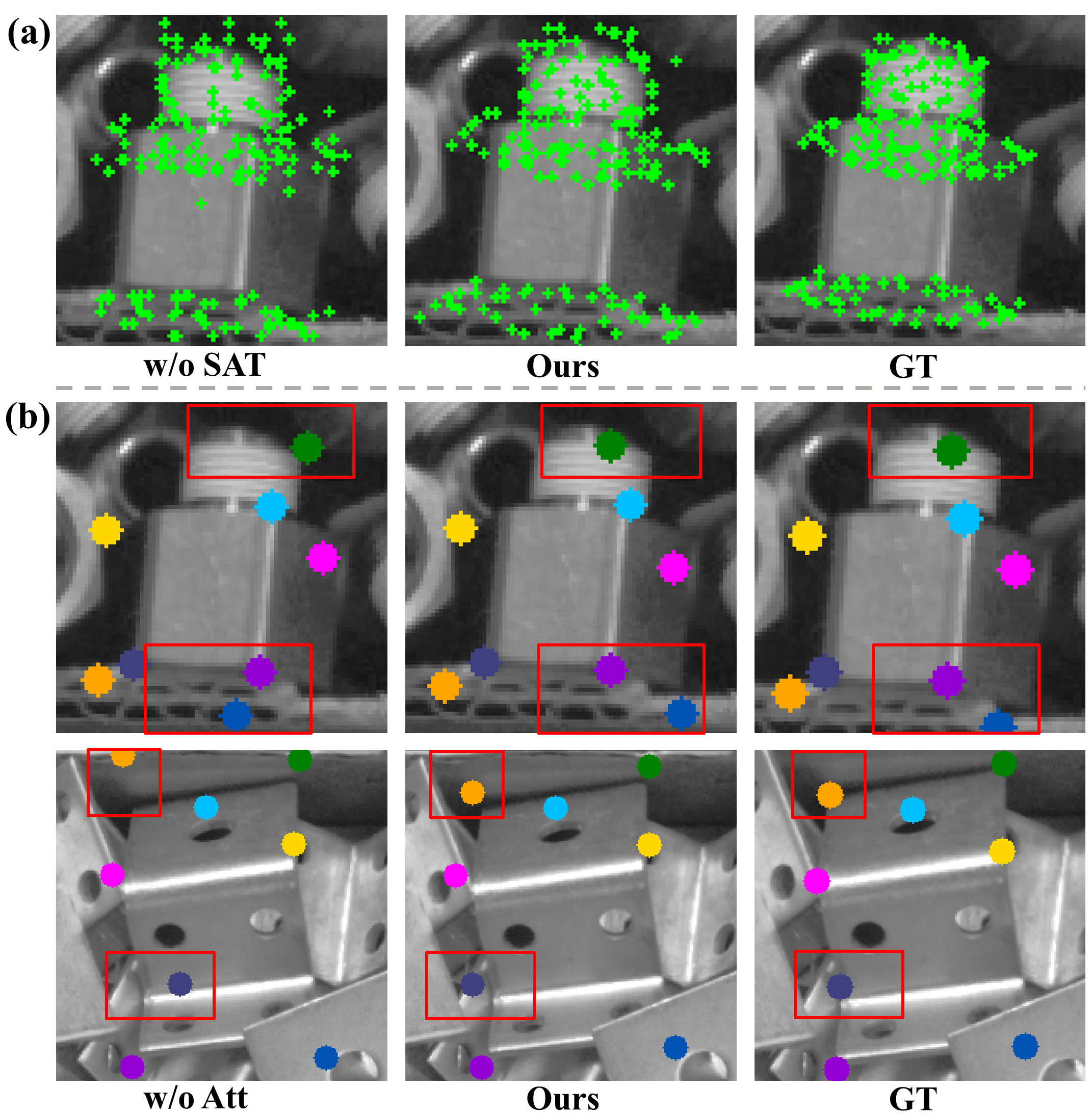}
    \label{fig:3}
    \vspace{-10pt}
    \begin{justify}
        \small Fig. 3: \textbf{(a)} Comparison of overall keypoint distributions with and without the SAT strategy. \textbf{(b)} Comparison of local keypoint localization accuracy with and without the Att. Please refer to the appendix~\ref{app:4} for additional results.
    \end{justify}
\end{figure}

\subsection{Ablation Studies}
Extensive ablation studies are conducted to evaluate each component, including SAT, attentional aggregation (Att) node update, keypoint density, and pose estimation method.
A stricter metric $(2\mathrm{mm}, 3^\circ)$ is introduced for more intuitive comparison of pose estimation performance.

\textbf{SAT strategy} As illustrated in Fig.~\ref{fig:3} (a), SAT strategy produces well-structured keypoint distributions with preserved geometric constraints, while its absence results in scattered and inconsistent predictions. Furthermore, the results presented in Table~\ref{tab:3} confirm that incorporating the SAT strategy significantly improves pose estimation performance.
The gains become more pronounced as the number of keypoints increases, since more accurate keypoint predictions enable more effective multi-view fusion.
Notably, the largest performance gain is achieved when the number of keypoints increases to 512, yielding improvements of 37.8\% and 50.3\% under the $(5\mathrm{mm}, 10^\circ)$ and $(2\mathrm{mm}, 3^\circ)$ metrics respectively.
This highlights the importance of the SAT strategy for robust multi-view dense keypoint fusion. 
\begin{figure}[!htbp]
	\centering
    \refstepcounter{figure}
	\includegraphics[width=\linewidth]{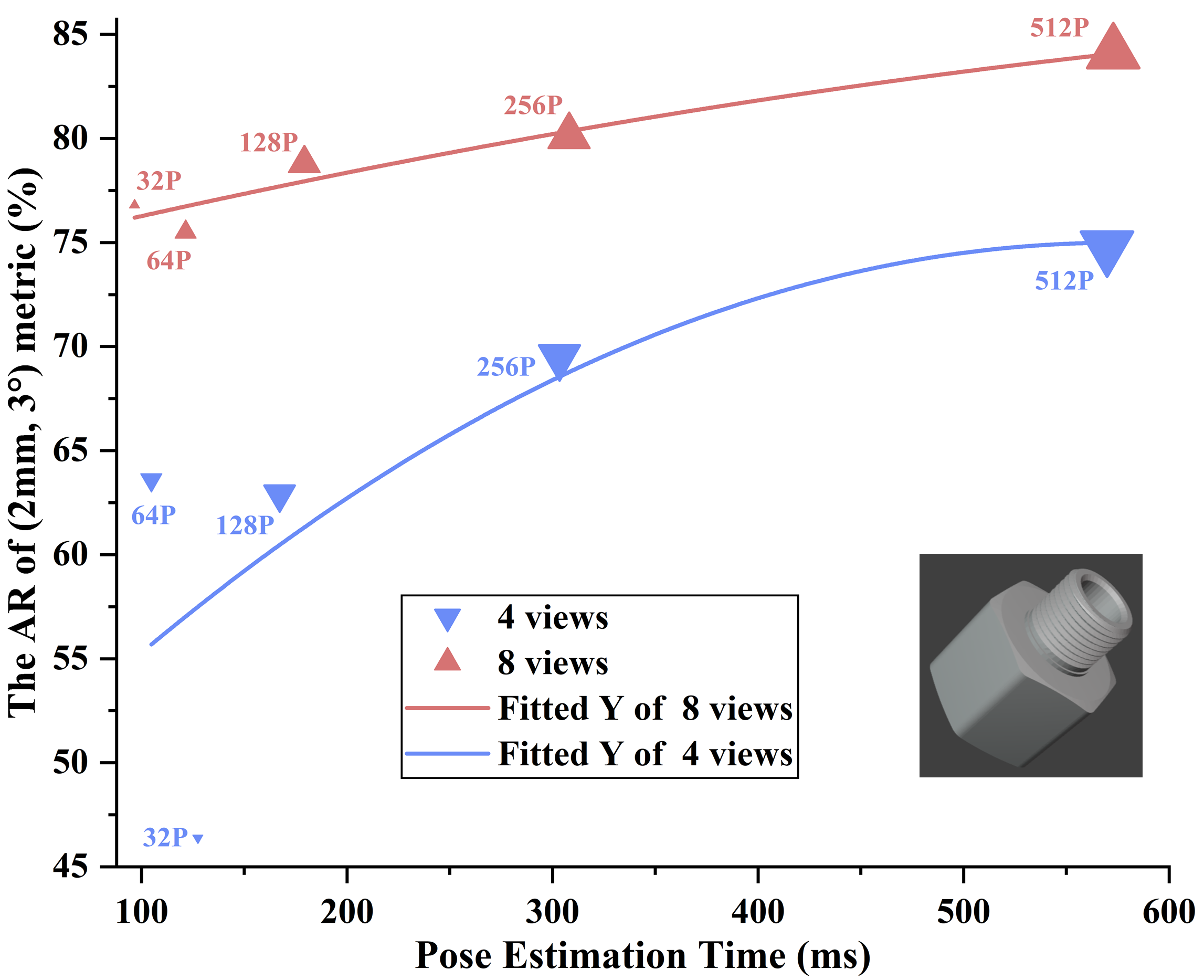}
    \label{fig:4}
    \vspace{-10pt}
    \begin{justify}
        \small Fig. 4: Pose estimation accuracy and runtime (including keypoint prediction and pose estimation) on the Tube-Fitting object under varying numbers of keypoints.
    \end{justify}
\end{figure}
\begin{figure*}[ht]
	\centering
    \refstepcounter{figure}
	\includegraphics[width=0.98\linewidth]{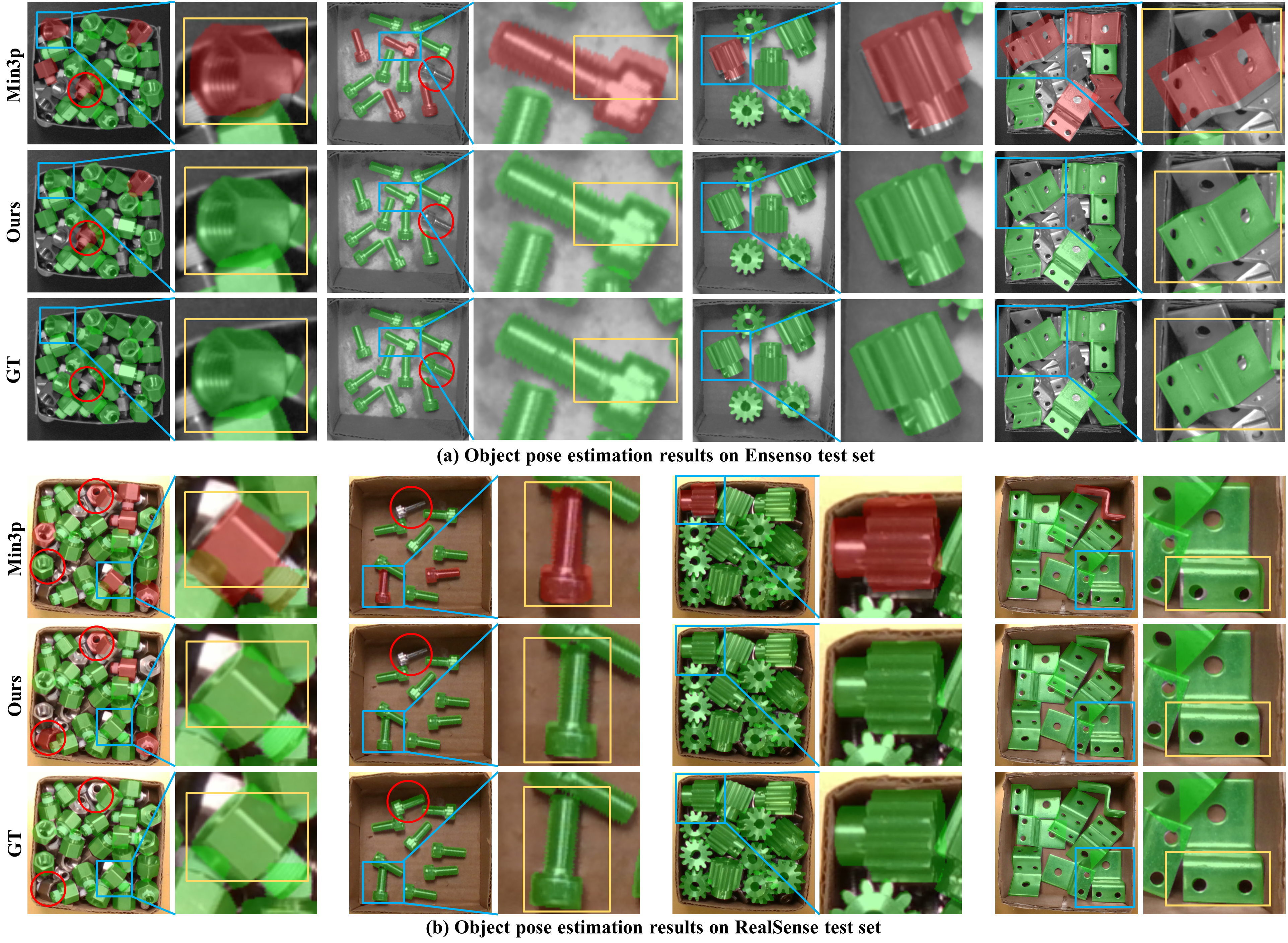}
    \label{fig:visual}
    \vspace{-8pt}
    \begin{justify}
        \small Fig. 5: Visualizations of the 6D pose estimation results on \textbf{(a)} Ensenso and  \textbf{(b)} RealSense test sets using 4-view input. The green or red renderings indicate whether the predicted poses satisfy the $(5\mathrm{mm}, 10^\circ)$ metric, while red circles highlight missed or falsely detected object. The enlarged views further highlight the fine-grained differences in pose estimation accuracy. More examples are in the appendix~\ref{app:5}.
    \end{justify}
\end{figure*}

\textbf{Attentional aggregation} As illustrated in Fig.~\ref{fig:3} (b), the original CheckerPose exhibits large deviations in the prediction of certain keypoints. In contrast, our keypoint network with Att node update more effectively captures the geometric constraints among neighboring dense keypoints, leading to more geometry-aware and accurate keypoint localization.
Meanwhile, the results presented in Table~\ref{tab:4} demonstrate that our Att node update significantly improves pose estimation performance on most objects.
Notably, under the $(2\mathrm{mm}, 3^\circ)$ metric, our method achieves improvements of 2.8\% and 3.6\% in the 4-view and 8-view settings respectively, confirming the effectiveness of our design.

\begin{table}[ht]
	\centering
    \refstepcounter{table}
    \renewcommand{\arraystretch}{1.09}
	\begin{tabular}{c|ccc|ccc}
		\hline
		\multirow{2}{*}{Point num} & \multicolumn{3}{c|}{(5, 10)}         & \multicolumn{3}{c}{(2, 3)}        \\ \cline{2-7}
		                           & w/o SAT    & w SAT      & $\Delta$   & w/o SAT   & w SAT     & $\Delta$  \\ \hline
		32                         & 46.4       & 82.8       & 36.4       & 16.6      & 46.4      & 29.8      \\
		64                         & 47.0       & 82.8       & 35.8       & 21.2      & 63.6      & 42.4      \\
		128                        & 54.3       & 88.1       & 33.8       & 26.5      & 62.9      & 36.4      \\
		256                        & 53.6       & 89.4       & 35.8       & \textbf{28.5}      & 69.5      & 41.0      \\
		512                        & \textbf{55.6}       & \textbf{93.4}       & \textbf{37.8}       & 24.5      & \textbf{74.8}      & \textbf{50.3}      \\ \hline
	\end{tabular}
    \vspace{0pt}
    \label{tab:3}
    \begin{justify}
        \small Table \uppercase\expandafter{\romannumeral3}: Comparison of pose estimation performance with and without the SAT strategy on the Tube-Fitting object from the Ensenso test set.
    \end{justify}
\end{table}

\begin{table}[ht]
	\centering
    \refstepcounter{table}
    \label{tab:4}
    \renewcommand{\arraystretch}{1.1}
	\begin{tabularx}{\columnwidth}{>{\centering\arraybackslash}p{2.0cm}|*{4}{X}|*{4}{X}}
		\hline
		\multirow{3}{*}{object} & \multicolumn{4}{c|}{4 views}                                      & \multicolumn{4}{c}{8 views}                                   \\ \cline{2-9}
		                        & \multicolumn{2}{c}{(5, 10)}     & \multicolumn{2}{c|}{(2, 3)}     & \multicolumn{2}{c}{(5, 10)}     & \multicolumn{2}{c}{(2, 3)}    \\ \cline{2-9}
		                        & Ori            & Att            & Ori            & Att            & Ori            & Att            & Ori           & Att           \\ \hline
		Tube-Fitting             & 87.4           & \textbf{89.4}  & 66.9           & \textbf{69.5}  & \textbf{98.7}  & 96.0           & \textbf{80.8} & 80.1          \\
		Chrome-Screw             & 88.2           & \textbf{92.9}  & 37.9           & \textbf{43.8}  & 96.4           & \textbf{97.6}  & 44.4          & \textbf{56.8} \\
		Eye-Bolt                 & 78.4           & \textbf{81.1}  & \textbf{36.5}  & 31.1           & \textbf{85.1}           & \textbf{85.1}           & \textbf{44.6}          & \textbf{44.6}          \\
		Gear                    & 92.6           & \textbf{92.9}  & 58.0           & \textbf{71.6}  & \textbf{96.3}           & \textbf{96.3}           & 76.5          & \textbf{82.7} \\
		Zigzag                  & 93.1           & \textbf{100.0} & 46.6           & \textbf{48.3}  & 96.6           & \textbf{100.0} & \textbf{63.8} & 58.6          \\
		DIN-Connector            & 86.5           & \textbf{89.7}  & 52.4           & \textbf{53.2}  & 85.7           & \textbf{92.9}  & 55.6          & \textbf{65.9} \\
		DSub-Connector          & \textbf{75.2}           & \textbf{75.2}           & 18.0           & \textbf{18.6}  & \textbf{81.4}  & 76.4           & 22.4          & \textbf{24.2} \\
		ALL                     & 85.9           & \textbf{88.7}  & 45.2           & \textbf{48.0}  & 91.5           & \textbf{92.0}  & 55.4          & \textbf{59.0} \\ \hline
	\end{tabularx}
    \vspace{-1pt}
    \begin{justify}
        \small Table \uppercase\expandafter{\romannumeral4}: Comparison of pose estimation performance between our Att node update and the original (Ori) CheckerPose on Ensenso test set.
    \end{justify}
\end{table}

\textbf{Number of keypoints} Fig.~\ref{fig:4} shows the pose accuracy under the strict $(2\mathrm{mm}, 3^\circ)$ metric and the runtime, tested on an Intel® Core™ Ultra 9 285K CPU and Nvidia 5090 GPU. 
As expected, pose estimation accuracy exhibits a generally increasing trend with more keypoints in both 4-view and 8-view settings, validating the effectiveness of the DKPMV.
The observed variations in accuracy and runtime at 32, 64, and 128 keypoints stem from insufficient keypoint density, where prediction noise disrupts the effectiveness of keypoint fusion and increases the required number of RANSAC iterations during optimization.
This further confirms the importance of using an appropriate number of dense keypoints for effective fusion.

\textbf{Pose estimation method} As shown in Table~\ref{tab:5}, we compare our pose estimation method with the minimal three-point solver (Min3P)~\cite{lee2015minimal} with RANSAC~\cite{fischler1981random}.
Our method consistently outperforms Min3P across both metrics, with especially notable gains under the strict $(2\mathrm{mm}, 3^\circ)$ metric, achieving improvements of 29.7\% and 42.3\% in the 4-view and 8-view settings, respectively.
These results demonstrate the effectiveness of our pose estimation pipeline.
Meanwhile, NO improves the performance of both methods, demonstrating its effectiveness.
Although Min3P requires only three points to compute the pose, it fails to fully leverage the information from multi-view keypoints, resulting in lower pose estimation accuracy.
In contrast, our method achieves high-precision pose estimation by performing only the first two stages of optimization. Fig.~\ref{fig:visual} illustrates the pose estimation visualizations for both our method and Min3P.

\begin{table}[ht]
	\centering
    \refstepcounter{table}
    \renewcommand{\arraystretch}{1.1}
	\begin{tabular}{cccccccc}
		\hline
		\multirow{2}{*}{} & \multirow{2}{*}{Min3p} & Ours       & \multirow{2}{*}{NO} & \multicolumn{2}{c}{4 views}     & \multicolumn{2}{c}{8 views}     \\ \cline{5-8}
		                  &                        & (w/o NO)   &                     & (5, 10)        & (2, 3)         & (5, 10)        & (2, 3)         \\ \hline
		1                 & \checkmark             &            &                     & 41.1           & 1.87           & 67.3           & 5.3            \\
		2                 & \checkmark             &            & \checkmark          & 73.2           & 18.3           & 82.0           & 17.3           \\
		3                 &                        & \checkmark &                     & \textbf{88.7}           & 47.2           & \textbf{92.0}           & \textbf{59.0}           \\
		4                 &                        & \checkmark & \checkmark          & \textbf{88.7}           & \textbf{48.0}           & \textbf{92.0}           & \textbf{59.0}           \\ \hline
	\end{tabular}
    \vspace{-1pt}
    \label{tab:5}
    \begin{justify}
        \small Table \uppercase\expandafter{\romannumeral5}: Comparing our method with Min3P~\cite{lee2015minimal} on Ensenso test set for all objects. We adopt RANSAC followed by nonlinear optimization (NO) for Min3P to suppress outliers and refine the estimated poses, thereby ensuring a fair comparison.
    \end{justify}
\end{table}

\section{CONCLUSIONS}

This paper proposes DKPMV, a novel multi-view RGB pipeline for 6D pose estimation of textureless objects, which achieves dense keypoint-level fusion without relying on depth input.
Leveraging three-stage progressive optimization, Att, and SAT, our method effectively captures multi-view geometric cues.
Extensive evaluations on the ROBI dataset confirm that our approach sets a new SOTA among RGB-based methods and even surpasses RGB-D methods in most scenarios.
Future work will focus on modeling dense keypoint uncertainty to enhance multi-view geometric fusion.



\renewcommand*{\bibfont}{\footnotesize}
\printbibliography

@inproceedings{stoiber2022iterative,
  title={Iterative corresponding geometry: Fusing region and depth for highly efficient 3d tracking of textureless objects},
  author={Stoiber, Manuel and Sundermeyer, Martin and Triebel, Rudolph},
  booktitle={Proceedings of the IEEE/CVF Conference on Computer Vision and Pattern Recognition (CVPR)},
  pages={6855--6865},
  year={2022}
}

@inproceedings{jin2023online,
  title={Online hand-eye calibration with decoupling by 3d textureless object tracking},
  author={Jin, Li and Xie, Kang and Chen, Wenxuan and Cao, Xin and Li, Yuehua and Li, Jiachen and Qian, Jiankai and Qin, Xueying},
  booktitle={Proceedings of 2023 IEEE International Conference on Robotics and Automation (ICRA)},
  pages={11453--11460},
  year={2023},
  organization={IEEE}
}

@inproceedings{yang2021robi,
  title={Robi: A multi-view dataset for reflective objects in robotic bin-picking},
  author={Yang, Jun and Gao, Yizhou and Li, Dong and Waslander, Steven L},
  booktitle={Proceedings of 2021 IEEE/RSJ International Conference on Intelligent Robots and Systems (IROS)},
  pages={9788--9795},
  year={2021},
  organization={IEEE}
}

@article{huang2025xyz,
  title={XYZ-IBD: High-precision Bin-picking Dataset for Object 6D Pose Estimation Capturing Real-world Industrial Complexity},
  author={Huang, Junwen and Liang, Jizhong and Hu, Jiaqi and Sundermeyer, Martin and Yu, Peter KT and Navab, Nassir and Busam, Benjamin},
  journal={arXiv preprint arXiv:2506.00599},
  year={2025}
}

@inproceedings{shi2024asgrasp,
  title={Asgrasp: Generalizable transparent object reconstruction and 6-dof grasp detection from rgb-d active stereo camera},
  author={Shi, Jun and Yong, A and Jin, Yixiang and Li, Dingzhe and Niu, Haoyu and Jin, Zhezhu and Wang, He},
  booktitle={Proceedings of 2024 IEEE international conference on robotics and automation (ICRA)},
  pages={5441--5447},
  year={2024},
  organization={IEEE}
}

@article{zhang2024instance,
  title={Instance-level 6D pose estimation based on multi-task parameter sharing for robotic grasping},
  author={Zhang, Liming and Zhou, Xin and Liu, Jiaqing and Wang, Can and Wu, Xinyu},
  journal={Scientific Reports},
  volume={14},
  number={1},
  pages={7801},
  year={2024},
  publisher={Nature Publishing Group UK London}
}

@inproceedings{kim2024sim,
  title={Sim-to-real Object Pose Estimation for Random Bin Picking},
  author={Kim, Boyoung and Min, Junhong},
  booktitle={Proceedings of 2024 IEEE International Conference on Robotics and Automation (ICRA)},
  pages={10749--10756},
  year={2024},
  organization={IEEE}
}

@article{chen2025zerobp,
  title={ZeroBP: Learning Position-Aware Correspondence for Zero-shot 6D Pose Estimation in Bin-Picking},
  author={Chen, Jianqiu and Zhou, Zikun and Li, Xin and Zheng, Ye and Bao, Tianpeng and He, Zhenyu},
  journal={arXiv preprint arXiv:2502.01004},
  year={2025}
}

@INPROCEEDINGS{9197461,
  author={Gao, Ge and Lauri, Mikko and Wang, Yulong and Hu, Xiaolin and Zhang, Jianwei and Frintrop, Simone},
  booktitle={2020 IEEE International Conference on Robotics and Automation (ICRA)}, 
  title={6D Object Pose Regression via Supervised Learning on Point Clouds}, 
  year={2020},
  volume={},
  number={},
  pages={3643-3649}
}

@inproceedings{gao2021cloudaae,
  title={Cloudaae: Learning 6d object pose regression with on-line data synthesis on point clouds},
  author={Gao, Ge and Lauri, Mikko and Hu, Xiaolin and Zhang, Jianwei and Frintrop, Simone},
  booktitle={Proceedings of 2021 IEEE International Conference on Robotics and Automation (ICRA)},
  pages={11081--11087},
  year={2021},
  organization={IEEE}
}

@inproceedings{cai2022ove6d,
  title={Ove6d: Object viewpoint encoding for depth-based 6d object pose estimation},
  author={Cai, Dingding and Heikkil{\"a}, Janne and Rahtu, Esa},
  booktitle={Proceedings of the IEEE/CVF Conference on Computer Vision and Pattern Recognition (CVPR)},
  pages={6803--6813},
  year={2022}
}

@inproceedings{wang2019densefusion,
  title={Densefusion: 6d object pose estimation by iterative dense fusion},
  author={Wang, Chen and Xu, Danfei and Zhu, Yuke and Mart{\'\i}n-Mart{\'\i}n, Roberto and Lu, Cewu and Fei-Fei, Li and Savarese, Silvio},
  booktitle={Proceedings of the IEEE/CVF Conference on Computer Vision and Pattern Recognition (CVPR)},
  pages={3343--3352},
  year={2019}
}

@INPROCEEDINGS{he2020pvn3ddeeppointwise3d,
  author={He, Yisheng and Sun, Wei and Huang, Haibin and Liu, Jianran and Fan, Haoqiang and Sun, Jian},
  booktitle={Proceedings of IEEE/CVF Conference on Computer Vision and Pattern Recognition (CVPR)}, 
  title={PVN3D: A Deep Point-Wise 3D Keypoints Voting Network for 6DoF Pose Estimation}, 
  year={2020},
  volume={},
  number={},
  pages={11629-11638}}

@article{saadi2021optimizing,
  title={Optimizing rgb-d fusion for accurate 6dof pose estimation},
  author={Saadi, Lounes and Besbes, Bassem and Kramm, Sebastien and Bensrhair, Abdelaziz},
  journal={IEEE Robotics and Automation Letters},
  volume={6},
  number={2},
  pages={2413--2420},
  year={2021},
  publisher={IEEE}
}

@ARTICLE{ni2025reasoninglearningperceptualmetric,
  author={Ni, Peiyuan and Chew, Chee Meng and Ang, Marcelo H. and Chirikjian, Gregory S.},
  journal={IEEE Robotics and Automation Letters}, 
  title={Reasoning and Learning a Perceptual Metric for Self-Training of Reflective Objects in Bin-Picking with a Low-cost Camera}, 
  year={2025},
  volume={},
  number={},
  pages={1-8}}

@inproceedings{yang2022next,
  title={Next-best-view prediction for active stereo cameras and highly reflective objects},
  author={Yang, Jun and Waslander, Steven L},
  booktitle={Proceedings of 2022 International Conference on Robotics and Automation (ICRA)},
  pages={3684--3690},
  year={2022},
  organization={IEEE}
}

@INPROCEEDINGS{junyang2023,
  author={Yang, Jun and Xue, Wenjie and Ghavidel, Sahar and Waslander, Steven L.},
  booktitle={Proceedings of 2023 IEEE International Conference on Robotics and Automation (ICRA)}, 
  title={6D Pose Estimation for Textureless Objects on RGB Frames using Multi-View Optimization}, 
  year={2023},
  pages={2905--2912}
}

@inproceedings{zhang2022transnet,
  title={Transnet: Category-level transparent object pose estimation},
  author={Zhang, Huijie and Opipari, Anthony and Chen, Xiaotong and Zhu, Jiyue and Yu, Zeren and Jenkins, Odest Chadwicke},
  booktitle={Proceedings of the European Conference on Computer Vision (ECCV)},
  pages={148--164},
  year={2022},
  organization={Springer}
}

@inproceedings{liu2020keypose,
  title={Keypose: Multi-view 3d labeling and keypoint estimation for transparent objects},
  author={Liu, Xingyu and Jonschkowski, Rico and Angelova, Anelia and Konolige, Kurt},
  booktitle={Proceedings of the IEEE/CVF Conference on Computer Vision and Pattern Recognition (CVPR)},
  pages={11602--11610},
  year={2020}
}

@INPROCEEDINGS{Chai2020DeepDepthFusion,
  author={Chai, Chun-Yu and Wu, Yu-Po and Tsao, Shiao-Li},
  booktitle={2020 IEEE International Conference on Robotics and Automation (ICRA)},
  title={Deep Depth Fusion for Black, Transparent, Reflective and Texture-Less Objects},
  year={2020},
  pages={6766-6772}
}

@inproceedings{cop2021new,
  title={New metrics for industrial depth sensors evaluation for precise robotic applications},
  author={Cop, Konrad P and Peters, Arne and {\v{Z}}agar, Bare L and Hettegger, Daniel and Knoll, Alois C},
  booktitle={Proceedings of 2021 IEEE/RSJ International Conference on Intelligent Robots and Systems (IROS)},
  pages={5350--5356},
  year={2021},
  organization={IEEE}
}

@article{liu2024deeplearningbasedobjectpose,
  title={Deep learning-based object pose estimation: A comprehensive survey},
  author={Liu, Jian and Sun, Wei and Yang, Hui and Zeng, Zhiwen and Liu, Chongpei and Zheng, Jin and Liu, Xingyu and Rahmani, Hossein and Sebe, Nicu and Mian, Ajmal},
  journal={arXiv preprint arXiv:2405.07801},
  year={2024}
}

@inproceedings{kleeberger2020single,
  title={Single shot 6d object pose estimation},
  author={Kleeberger, Kilian and Huber, Marco F},
  booktitle={Proceedings of 2020 IEEE International Conference on Robotics and Automation (ICRA)},
  pages={6239--6245},
  year={2020},
  organization={IEEE}
}

@inproceedings{su2022zebrapose,
  title={Zebrapose: Coarse to fine surface encoding for 6dof object pose estimation},
  author={Su, Yongzhi and Saleh, Mahdi and Fetzer, Torben and Rambach, Jason and Navab, Nassir and Busam, Benjamin and Stricker, Didier and Tombari, Federico},
  booktitle={Proceedings of the IEEE/CVF Conference on Computer Vision and Pattern Recognition (CVPR)},
  pages={6738--6748},
  year={2022}
}

@inproceedings{lian2023checkerpose,
  title={Checkerpose: Progressive dense keypoint localization for object pose estimation with graph neural network},
  author={Lian, Ruyi and Ling, Haibin},
  booktitle={Proceedings of the IEEE/CVF Conference on Computer Vision and Pattern Recognition (CVPR)},
  pages={14022--14033},
  year={2023}
}

@article{bauer2024challenges,
  title={Challenges for monocular 6d object pose estimation in robotics},
  author={Bauer, Dominik and H{\"o}nig, Peter and Weibel, Jean-Baptiste and Garc{\'\i}a-Rodr{\'\i}guez, Jos{\'e} and Vincze, Markus and others},
  journal={IEEE Transactions on Robotics},
  year={2024},
  publisher={IEEE}
}

@inproceedings{fu2021multi,
  title={A multi-hypothesis approach to pose ambiguity in object-based slam},
  author={Fu, Jiahui and Huang, Qiangqiang and Doherty, Kevin and Wang, Yue and Leonard, John J},
  booktitle={Proceedings of 2021 IEEE/RSJ International Conference on Intelligent Robots and Systems (IROS)},
  pages={7639--7646},
  year={2021},
  organization={IEEE}
}

@inproceedings{merrill2022symmetry,
  title={Symmetry and uncertainty-aware object slam for 6dof object pose estimation},
  author={Merrill, Nathaniel and Guo, Yuliang and Zuo, Xingxing and Huang, Xinyu and Leutenegger, Stefan and Peng, Xi and Ren, Liu and Huang, Guoquan},
  booktitle={Proceedings of the IEEE/CVF Conference on Computer Vision and Pattern Recognition (CVPR)},
  pages={14901--14910},
  year={2022}
}

@article{maninis2022vid2cad,
  title={Vid2cad: Cad model alignment using multi-view constraints from videos},
  author={Maninis, Kevis-Kokitsi and Popov, Stefan and Nie{\ss}ner, Matthias and Ferrari, Vittorio},
  journal={IEEE transactions on Pattern Analysis and Machine Intelligence},
  volume={45},
  number={1},
  pages={1320--1327},
  year={2022},
  publisher={IEEE}
}

@inproceedings{labbe2020cosypose,
  title={Cosypose: Consistent multi-view multi-object 6d pose estimation},
  author={Labb{\'e}, Yann and Carpentier, Justin and Aubry, Mathieu and Sivic, Josef},
  booktitle={Proceedings of the European Conference on Computer Vision (ECCV)},
  pages={574--591},
  year={2020},
  organization={Springer}
}

@article{shugurov2021multi,
  title={Multi-view object pose refinement with differentiable renderer},
  author={Shugurov, Ivan and Pavlov, Ivan and Zakharov, Sergey and Ilic, Slobodan},
  journal={IEEE Robotics and Automation Letters},
  volume={6},
  number={2},
  pages={2579--2586},
  year={2021},
  publisher={IEEE}
}

@INPROCEEDINGS{LiMVKP,
  author={Li, Alan and Schoellig, Angela P.},
  booktitle={Proceedings of 2023 IEEE International Conference on Robotics and Automation (ICRA)}, 
  title={Multi-View Keypoints for Reliable 6D Object Pose Estimation}, 
  year={2023},
  pages={6988--6994}
  }

@article{yang2025active,
  title={Active 6D Pose Estimation for Textureless Objects using Multi-View RGB Frames},
  author={Yang, Jun and Xue, Wenjie and Ghavidel, Sahar and Waslander, Steven L},
  journal={arXiv preprint arXiv:2503.03726},
  year={2025}
}

@inproceedings{choudhury2023tempo,
  title={Tempo: Efficient multi-view pose estimation, tracking, and forecasting},
  author={Choudhury, Rohan and Kitani, Kris M and Jeni, L{\'a}szl{\'o} A},
  booktitle={Proceedings of the IEEE/CVF Conference on Computer Vision and Pattern Recognition (CVPR)},
  pages={14750--14760},
  year={2023}
}

@inproceedings{pitteri2019object,
  title={On object symmetries and 6d pose estimation from images},
  author={Pitteri, Giorgia and Ramamonjisoa, Micha{\"e}l and Ilic, Slobodan and Lepetit, Vincent},
  booktitle={Proceedings of 2019 International Conference on 3D Vision (3DV)},
  pages={614--622},
  year={2019},
  organization={IEEE}
}

@article{umeyama2002least,
  title={Least-squares estimation of transformation parameters between two point patterns},
  author={Umeyama, Shinji},
  journal={IEEE Transactions on Pattern Analysis and Machine Intelligence},
  volume={13},
  number={4},
  pages={376--380},
  year={2002},
  publisher={IEEE}
}

@article{deng2021poserbpf,
  title={PoseRBPF: A Rao--Blackwellized particle filter for 6-D object pose tracking},
  author={Deng, Xinke and Mousavian, Arsalan and Xiang, Yu and Xia, Fei and Bretl, Timothy and Fox, Dieter},
  journal={IEEE Transactions on Robotics},
  volume={37},
  number={5},
  pages={1328--1342},
  year={2021},
  publisher={IEEE}
}

@inproceedings{sundermeyer2018implicit,
  title={Implicit 3d orientation learning for 6d object detection from rgb images},
  author={Sundermeyer, Martin and Marton, Zoltan-Csaba and Durner, Maximilian and Brucker, Manuel and Triebel, Rudolph},
  booktitle={Proceedings of the European Conference on Computer Vision (ECCV)},
  pages={699--715},
  year={2018}
}

@inproceedings{li2020pose,
  title={Pose-guided auto-encoder and feature-based refinement for 6-dof object pose regression},
  author={Li, Zhigang and Ji, Xiangyang},
  booktitle={Proceedings of 2020 IEEE International Conference on Robotics and Automation (ICRA)},
  pages={8397--8403},
  year={2020},
  organization={IEEE}
}

@inproceedings{xu20246d,
  title={6d-diff: A keypoint diffusion framework for 6d object pose estimation},
  author={Xu, Li and Qu, Haoxuan and Cai, Yujun and Liu, Jun},
  booktitle={Proceedings of the IEEE/CVF Conference on Computer Vision and Pattern Recognition (CVPR)},
  pages={9676--9686},
  year={2024}
}

@inproceedings{liu2021kdfnet,
  title={Kdfnet: Learning keypoint distance field for 6d object pose estimation},
  author={Liu, Xingyu and Iwase, Shun and Kitani, Kris M},
  booktitle={Proceedings of 2021 IEEE/RSJ International Conference on Intelligent Robots and Systems (IROS)},
  pages={4631--4638},
  year={2021},
  organization={IEEE}
}

@inproceedings{hu2020single,
  title={Single-stage 6d object pose estimation},
  author={Hu, Yinlin and Fua, Pascal and Wang, Wei and Salzmann, Mathieu},
  booktitle={Proceedings of the IEEE/CVF Conference on Computer Vision and Pattern Recognition (CVPR)},
  pages={2930--2939},
  year={2020}
}

@inproceedings{wang2021gdr,
  title={Gdr-net: Geometry-guided direct regression network for monocular 6d object pose estimation},
  author={Wang, Gu and Manhardt, Fabian and Tombari, Federico and Ji, Xiangyang},
  booktitle={Proceedings of the IEEE/CVF Conference on Computer Vision and Pattern Recognition (CVPR)},
  pages={16611--16621},
  year={2021}
}

@inproceedings{pavlakos20176,
  title={6-dof object pose from semantic keypoints},
  author={Pavlakos, Georgios and Zhou, Xiaowei and Chan, Aaron and Derpanis, Konstantinos G and Daniilidis, Kostas},
  booktitle={Proceedings of 2017 IEEE international Conference on Robotics and Automation (ICRA)},
  pages={2011--2018},
  year={2017},
  organization={IEEE}
}

@inproceedings{liu2021mfpn,
  title={MFPN-6D: Real-time one-stage pose estimation of objects on RGB images},
  author={Liu, Penglei and Zhang, Qieshi and Zhang, Jin and Wang, Fei and Cheng, Jun},
  booktitle={Proceedings of 2021 IEEE International Conference on Robotics and Automation (ICRA)},
  pages={12939--12945},
  year={2021},
  organization={IEEE}
}

@inproceedings{chen2022epro,
  title={Epro-pnp: Generalized end-to-end probabilistic perspective-n-points for monocular object pose estimation},
  author={Chen, Hansheng and Wang, Pichao and Wang, Fan and Tian, Wei and Xiong, Lu and Li, Hao},
  booktitle={Proceedings of the IEEE/CVF Conference on Computer Vision and Rattern Recognition (CVPR)},
  pages={2781--2790},
  year={2022}
}

@inproceedings{di2021so,
  title={So-pose: Exploiting self-occlusion for direct 6d pose estimation},
  author={Di, Yan and Manhardt, Fabian and Wang, Gu and Ji, Xiangyang and Navab, Nassir and Tombari, Federico},
  booktitle={Proceedings of the IEEE/CVF International Conference on Computer Vision (ICCV)},
  pages={12396--12405},
  year={2021}
}

@inproceedings{sarlin2020superglue,
  title={Superglue: Learning feature matching with graph neural networks},
  author={Sarlin, Paul-Edouard and DeTone, Daniel and Malisiewicz, Tomasz and Rabinovich, Andrew},
  booktitle={Proceedings of the IEEE/CVF Conference on Computer Vision and Pattern Recognition (CVPR)},
  pages={4938--4947},
  year={2020}
}

@article{wang2019dynamic,
  title={Dynamic graph cnn for learning on point clouds},
  author={Wang, Yue and Sun, Yongbin and Liu, Ziwei and Sarma, Sanjay E and Bronstein, Michael M and Solomon, Justin M},
  journal={ACM Transactions on Graphics (TOG)},
  volume={38},
  number={5},
  pages={1--12},
  year={2019},
  publisher={Acm New York, NY, USA}
}

@article{lee2015minimal,
  title={Minimal solutions for the multi-camera pose estimation problem},
  author={Lee, Gim Hee and Li, Bo and Pollefeys, Marc and Fraundorfer, Friedrich},
  journal={The International Journal of Robotics Research},
  volume={34},
  number={7},
  pages={837--848},
  year={2015},
  publisher={SAGE Publications Sage UK: London, England}
}

@article{fischler1981random,
  title={Random sample consensus: a paradigm for model fitting with applications to image analysis and automated cartography},
  author={Fischler, Martin A and Bolles, Robert C},
  journal={Communications of the ACM},
  volume={24},
  number={6},
  pages={381--395},
  year={1981},
  publisher={ACM New York, NY, USA}
}
\clearpage
\appendix

\subsection{Multi-view Instance Keypoints Matching}
\label{app:1}
Prior to dense keypoint fusion across views, it is crucial to establish accurate correspondences between predicted keypoints of the same object instance from different viewpoints.

During the association stage, we first select a pair of views with minimal baseline distance, denoted as $ I_{u} $ and $ I_{v} $. 
Suppose view $ I_{u} $ contains $ \alpha $ predicted instance keypoint sets $ \lbrace P_{u}^{i} \rbrace_{i=1}^{\alpha} $ and view $ I_{v} $ contains $ \beta $ sets $ \lbrace Q_{u}^{j} \rbrace_{j=1}^{\beta} $, where each $P_u^i, Q_v^j \in \mathbb{R}^{N \times 2}$ represents a dense 2D keypoint prediction of $N$ points. 
We aim to identify all matching pairs $(P_u^i, Q_v^j)$, where the correspondence between keypoints is established based on their semantic consistency.

Specifically, we enforce the epipolar constraint by defining the distance between $P_u^i$ and $Q_v^j$ as the mean Sampson distance over all semantically aligned keypoints:
\begin{equation}
D\left(i, j\right)=\frac{1}{\tilde{N}} \sum_{k=1}^{\tilde{N}} \frac{\left(\left(\bm{p}_{v}^{k}\right)^{\top} F \bm{q}_{u}^{k}\right)^{2}}{\left(F \bm{p}_{u}^{k}\right)_{1}^{2}+\left(F \bm{p}_{u}^{k}\right)_{2}^{2}+\left(F^{\top} \bm{q}_{v}^{k}\right)_{1}^{2}+\left(F^{\top} \bm{q}_{v}^{k}\right)_{2}^{2}}
\end{equation}
where $ \bm{p}_{u}^{k}\in P_{u}^{i} $ and $ \bm{q}_{v}^{k}\in Q_{v}^{j} $ denote the $k$-th semantically corresponding keypoints, and $ F\in \mathbb{R}^{3 \times 3} $ is the fundamental matrix between views $ I_{u} $ and $ I_{v} $.
Here, $ \tilde{N} $ denotes the number of keypoints that are simultaneously visible in both views, as indicated by the visibility code $ b_{v}=1 $, satisfying $ \tilde{N}<=N $.

We compute the Sampson distance $ D\left(i, j\right) $ for all pairs $(P_u^i, Q_v^j)$, where $ i\in\lbrace 1,...,\alpha \rbrace $ and $ j\in\lbrace 1,...,\beta \rbrace $.
A pair $ (\bm{i}^{*},\bm{j}^{*}) $ is considered a valid match if and only if it satisfies the mutual nearest constraint:
\begin{equation}
D\left(\bm{i}^{*}, \bm{j}^{*}\right)=\min _{j} D\left(\bm{i}^{*}, j\right),\quad D\left(\bm{i}^{*}, \bm{j}^{*}\right)=\min _{i} D\left(i, \bm{j}^{*}\right)
\end{equation}

After establishing correspondences between the first two views, the matched keypoints are triangulated using known camera extrinsics, and additional correspondences in the remaining views are determined by selecting keypoints that minimize the average reprojection error of the resulting 3D points.

\begin{figure*}[ht]
	\centering
    \refstepcounter{figure}
	\includegraphics[width=\linewidth]{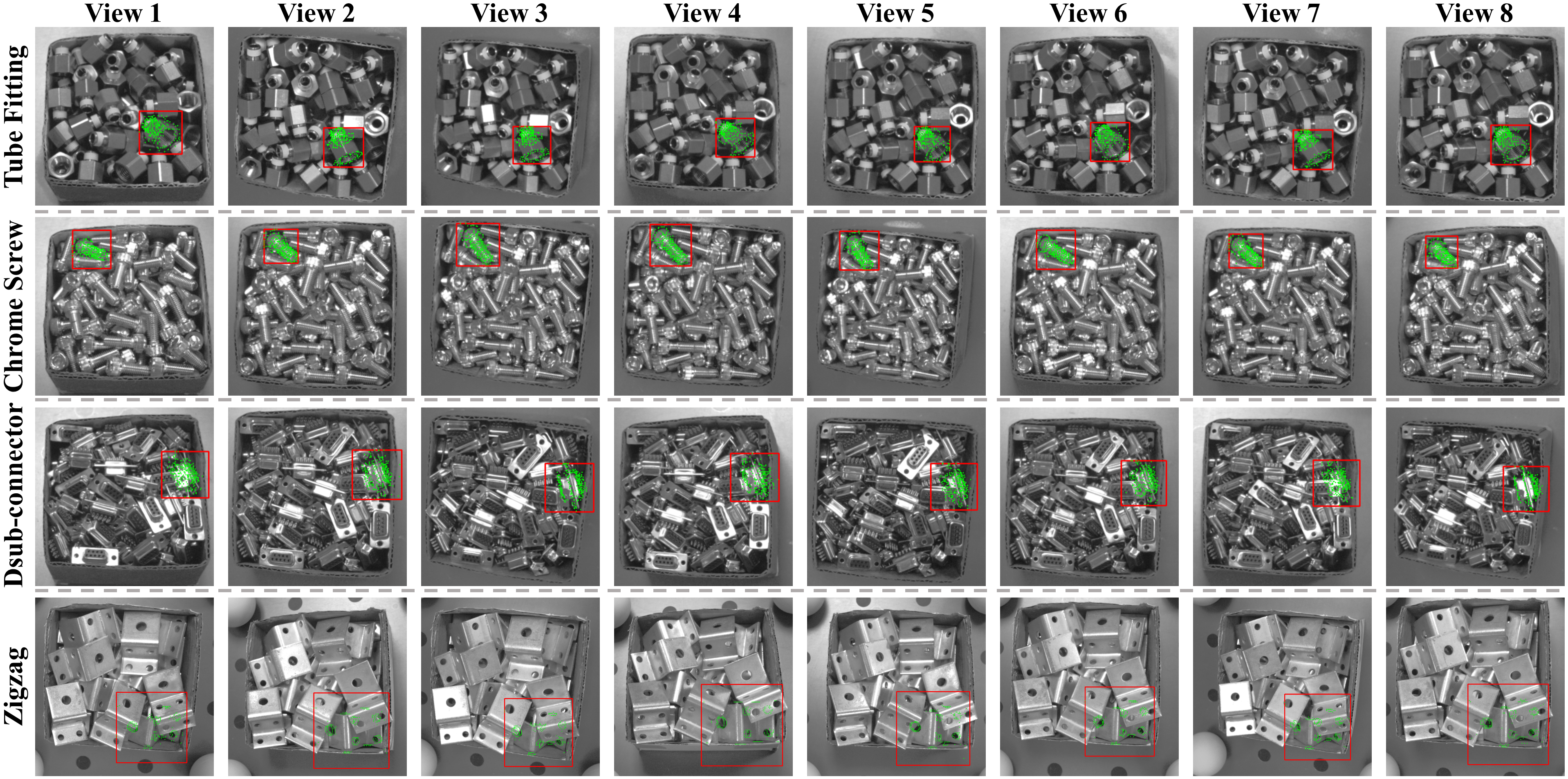}
    \label{fig:6}
    \vspace{-12pt}
    \begin{justify}
        \small Fig. 6: Visualization of multi-view keypoint matching. The results from 8 different views are presented to demonstrate the effectiveness of our matching strategy.
    \end{justify}
\end{figure*}

As shown in Fig.~\ref{fig:6}, our dense keypoint strategy achieves accurate multi-view matching even in cluttered scenes with densely stacked objects. Moreover, it demonstrates strong robustness to occlusions, as evidenced by the DSub-connector and Zigzag objects in the last two rows.

\subsection{Maximum Likelihood Estimation of Pose Estimation}
\label{app:2}
We define the reprojection error between the predicted keypoint $\bm{\tilde{p}}_{i}^{j}$ and its theoretical projection $\bm{p}_{i}^{j}$ as $\bm{e}_{i}^{j} = \bm{\tilde{p}}_{i}^{j} - \bm{p}_{i}^{j}$.

Assuming that all keypoints are independently and identically distributed (i.i.d.) according to an isotropic Gaussian distribution with equal variance, the probability density function of  $\bm{e}_{i}^{j}$ is given by:
\begin{equation}
p\left(\boldsymbol{e}_{i}^{j} \mid \mathbf{P}_{i}, \mathbf{R}, \mathbf{t}, \mathbf{R}_{j}, \mathbf{t}_{j}, \mathbf{K}_{j}\right)=\frac{1}{2 \pi \sigma^{2}} \exp \left(-\frac{1}{2 \sigma^{2}}\left\|\boldsymbol{e}_{i}^{j}\right\|_{2}\right) \label{eq:14}
\end{equation}
By omitting the known quantities $\mathbf{K}_{j}$, $\mathbf{R}_{j}$, $\mathbf{t}_{j}$, and $\mathbf{P}_{i}$, Eq.~\ref{eq:14} can be simplified as:
\begin{equation}
p\left(\boldsymbol{e}_{i}^{j} \mid \mathbf{R}, \boldsymbol{t}\right)=\frac{1}{2 \pi \sigma^{2}} \exp \left(-\frac{1}{2 \sigma^{2}}\left\|\boldsymbol{e}_{i}^{j}\right\|_{2}\right) \label{eq:15}
\end{equation}
Similarly, the posterior distribution of the object pose can be denoted as $ p\left(\mathbf{R}, \boldsymbol{t}\mid \boldsymbol{e}_{i}^{j} \right) $.
According to Bayes’ theorem, it follows that:
\begin{equation}
p\left(\mathbf{R}, \boldsymbol{t}\mid \boldsymbol{e}_{i}^{j} \right)\propto p\left(\boldsymbol{e}_{i}^{j} \mid \mathbf{R}, \boldsymbol{t}\right)p(\mathbf{R}, \boldsymbol{t})
\end{equation}
Assuming that the keypoint observation noise is independently distributed and the prior $p(\mathbf{R}, \boldsymbol{t})$ is uniform, we derive the following formulation in conjunction with Eq.~\ref{eq:15}:
\begin{equation}
p\left(\mathbf{R}, \boldsymbol{t}\mid \boldsymbol{e}_{i}^{j}\right)\infty\prod_{i=1}^{\tilde{N}} \prod_{j=1}^{V} \frac{1}{2 \pi \sigma^{2}} \exp \left(-\frac{1}{2 \sigma^{2}}\left\|e_{i}^{j}\right\|_{2}\right) \label{eq:17}
\end{equation}
Eq.~\ref{eq:17} defines the likelihood function for multi-view pose estimation.
By applying the logarithm, i.e., $ log(\cdot) $, and performing maximum likelihood estimation, we obtain the optimal estimate of the object pose as:
\begin{equation}
\mathbf{R}^{*}, \mathbf{t}^{*}=\underset{\mathbf{R}, \boldsymbol{t}}{\arg \min } \sum_{i=1}^{\tilde{N}} \sum_{j=1}^{V}\left\|\tilde{\boldsymbol{p}}_{i}^{j}-\boldsymbol{p}_{i}^{j}\right\|_{2}
\end{equation}

\subsection{Implementation details of three-stage progressive pose estimation}
\label{app:3}
\textbf{Optimized dense point cloud generation} For each RANSAC iteration, two views ($ \mathcal{I}_u,\mathcal{I}_\omega $) are randomly sampled.
For each keypoint $ i $, an initial 3D estimate $ \mathbf{\tilde{P}}_{i}^{(0)} $ is triangulated by solving the multi-view reprojection constraint:
\begin{equation}
\mathbf{\tilde{p}}_{i}^{j} \times\left(\mathbf{K}_{j}\left[\mathbf{R}_{j} \mid \mathbf{t}_{j}\right] \mathbf{\tilde{P}}_{i}^{(0)}\right)=\mathbf{0}, \quad j \in \lbrace u, \omega\rbrace \label{eq:18}
\end{equation}
These 3D points are then reprojected to all views using the camera model:
\begin{equation}
\mathbf{\hat{p}}_{i}^{j}=\pi(\mathbf{K}_{j},\mathbf{R}_{j},\mathbf{t}_{j},\mathbf{\tilde{P}}_{i}^{(0)}), \quad j\in \lbrace 1,\ldots,V \rbrace
\end{equation}
where $ \pi(\cdot) $ denotes the projection function. 

For each point, the set of inlier views is selected via reprojection error below a threshold $ \tau_{1} $:
\begin{equation}
\mathcal{V}_{i}=\left\{v \mid\left\|\mathbf{\hat{p}}_{i}^{v}-\mathbf{\tilde{p}}_{i}^{v}\right\|_{2}<\tau_{1}\right\}
\end{equation}
and the total number of inliers for the current hypothesis is computed as:
\begin{equation}
N_{\text {inliers }}=\sum_{i=1}^{\tilde{N}} \sum_{v=1}^{V} \mathbbm{1}\left(\left\|\mathbf
{\hat{p}}_{i}^{v}-\mathbf{\tilde{p}}_{i}^{v}\right\|_{2}<\tau_{1}\right)
\end{equation}
where $ \mathbbm{1}(\cdot) $ is an indicator function.

Meanwhile, we define a scoring function to evaluate and select the optimal RANSAC result, which is formally defined as follows:
\begin{equation}
\mathrm{Score}=\dfrac{N_{\text {inliers }}}{1+\sum_{i=1}^{\tilde{N}}\sum_{v=1}^{V}\left\|\mathbf
{\hat{p}}_{i}^{v}-\mathbf{\tilde{p}}_{i}^{v}\right\|_{2}}
\end{equation}
The candidate with the highest $\mathrm{Score}$ is selected as the optimal result. 
Based on Eq.~\ref{eq:18}, we then construct an overdetermined system by selecting $J \in \mathcal{V}_i$, from which the global solution $\mathbf{\tilde{P}}_{i}^{*}$ is computed via SVD.

\textbf{Pose alignment with dense correspondence} After reconstructing the optimal dense point cloud $ \mathcal{P}=\lbrace \mathbf{\tilde{P}}_{i}^{*}\in \mathbb{R}^{3}\rbrace_{i=1}^{\tilde{N}} $, we estimate the initial object pose by aligning it to the reference 3D keypoints $ \mathcal{P}_{o}=\lbrace \mathbf{P}_{i}\in \mathbb{R}^{3}\rbrace_{i=1}^{\tilde{N}} $ defined on the CAD model.
Based on Eq.~\ref{eq:10}, we estimate pose ($ \mathbf{R}^{*}, \mathbf{t}^{*} $) by employing the Umeyama algorithm~\cite{umeyama2002least}. 

RANSAC is then employed to robustly remove outliers by randomly selecting three points $ \mathbf{\tilde{P}}_{i}^{*}\in \mathcal{P} $ to compute an initial pose ($ \mathbf{R}_{o},\mathbf{t}_{o} $).
The number of inliers is defined as the count of correspondences where the Euclidean distance between the transformed source target points falls below a threshold $ \tau_{2} $, i.e.,
\begin{equation}
N_{\text{inlier}}^{3d}=\sum_{i=1}^{\tilde{N}} \mathbbm{1}\left(\left\|\mathbf{R}_{o} \mathbf{P}_{i}+\mathbf{t}_{o}-\mathbf{\tilde{P}}_{i}^{*}\right\|_{2}<\tau_{2}\right)
\end{equation}
We select the hypothesis with the highest inlier count and compute the final global solution using these inliers via Eq.~\ref{eq:10}.



\subsection{Qualitative Results of Keypoint Prediction}
\label{app:4}
\begin{figure*}[ht]
	\centering
    \refstepcounter{figure}
	\includegraphics[width=0.95\linewidth]{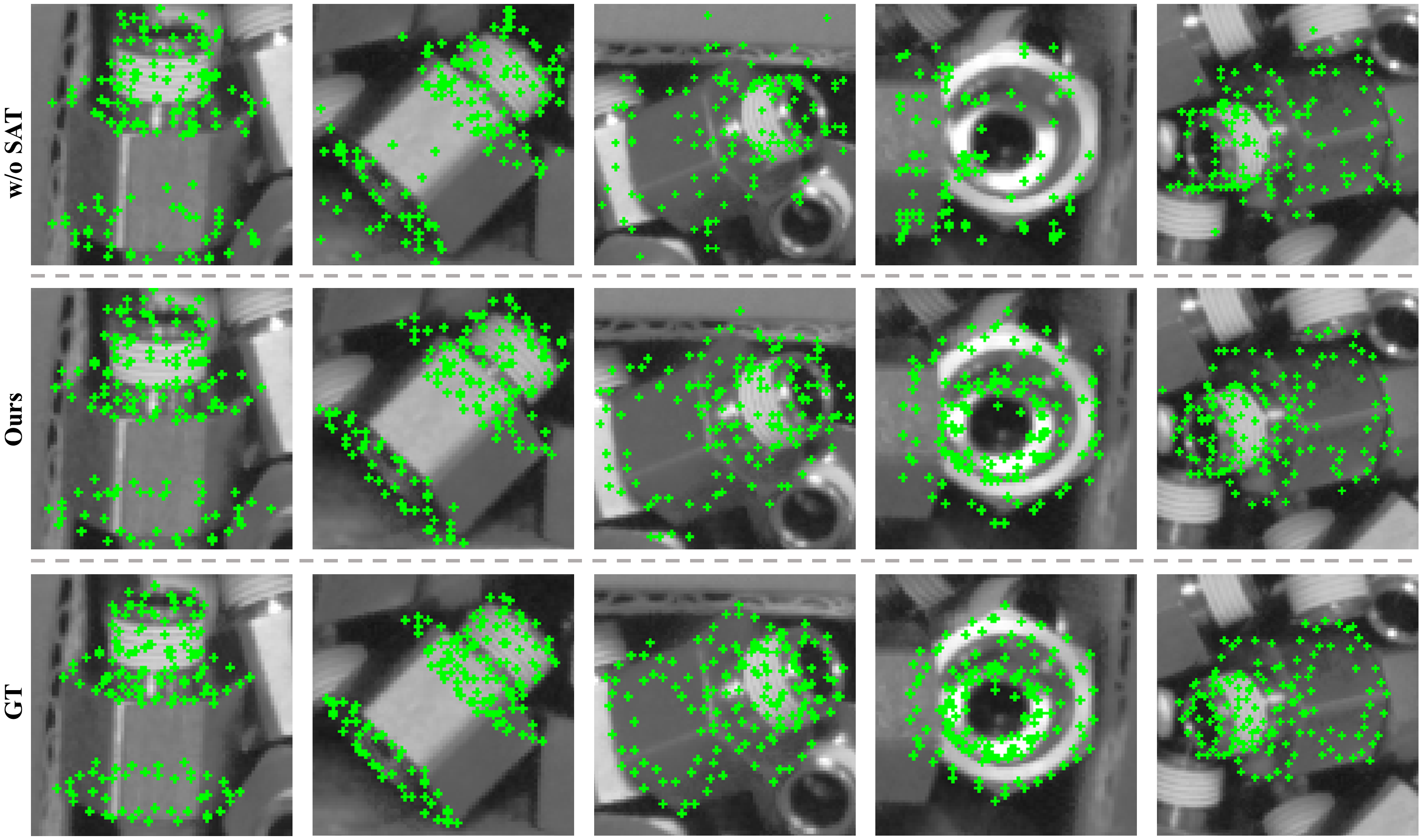}
    \label{fig:7}
    \vspace{-3pt}
    \begin{justify}
        \small Fig. 7: Comparison of dense keypoint prediction distributions with and without the SAT strategy. The absence of SAT leads to scattered dense keypoint predictions, preventing the network from effectively modeling the geometric structure among keypoints.
    \end{justify}
\end{figure*}

\begin{figure*}[ht]
	\centering
    \refstepcounter{figure}
	\includegraphics[width=0.95\linewidth]{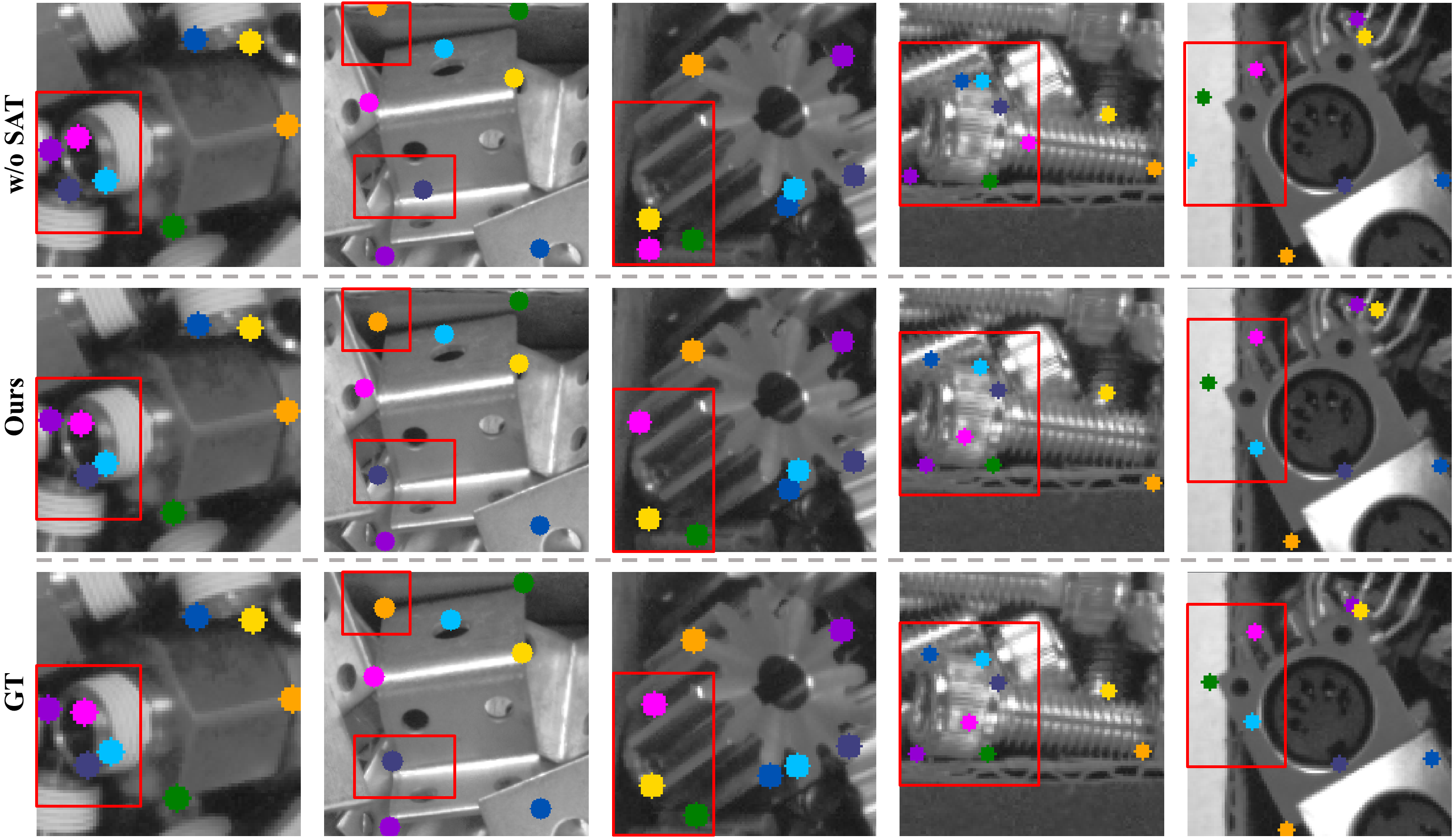}
    \label{fig:8}
    \vspace{-3pt}
    \begin{justify}
        \small Fig. 8: Comparison of local keypoint localization across different objects with and without the Att. Our method accurately captures the geometric relationships between neighboring keypoints, enabling more precise keypoint localization.
    \end{justify}
\end{figure*}

Figs.~\ref{fig:7} and \ref{fig:8} present additional qualitative comparisons. Fig.~\ref{fig:7} focuses on the impact of the SAT strategy on dense keypoint prediction distributions, while Fig.~\ref{fig:8} highlights the effect of the Att module on local keypoint localization across various objects.

\subsection{Qualitative Results of Pose Estimation}
\label{app:5}
\begin{figure*}[!ht]
	\centering
    \refstepcounter{figure}
	\includegraphics[width=\linewidth]{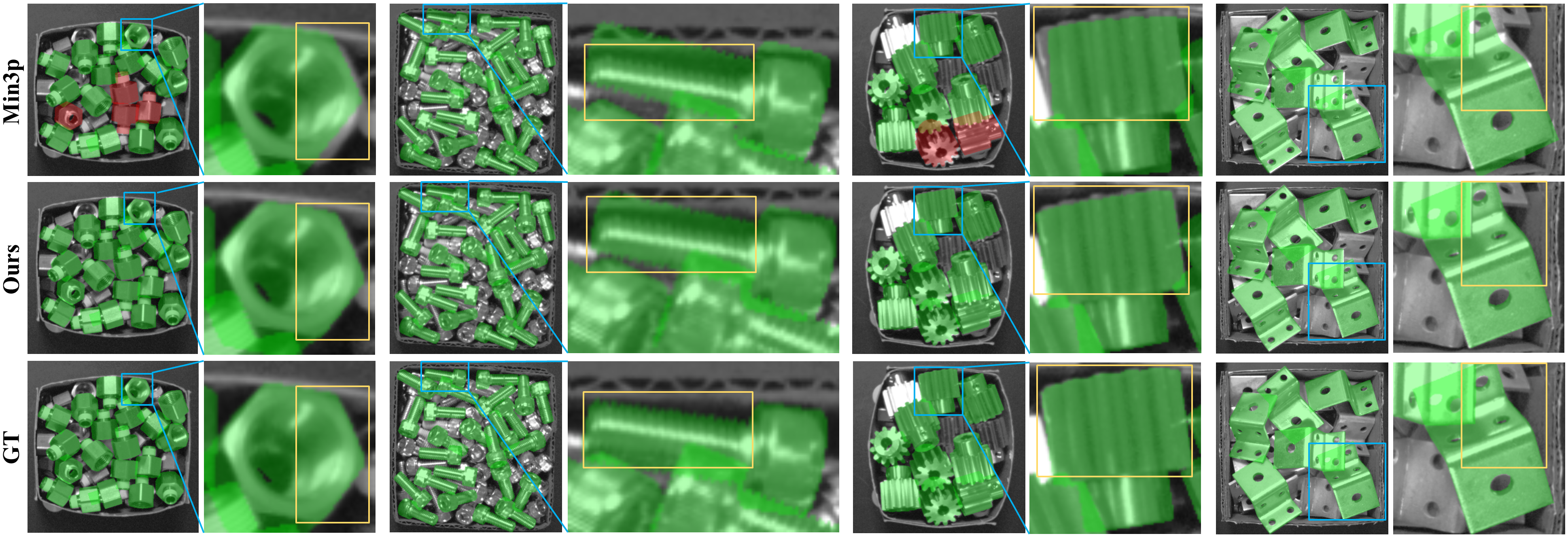}
    \label{fig:9}
    \vspace{-12pt}
    \begin{justify}
        \small Fig. 9: Visualizations of the 6D pose estimation results on Ensenso test set using 8-view input.
    \end{justify}
\end{figure*}

As shown in Fig.~\ref{fig:9}, increasing the number of views to 8 improves the accuracy of Min3P on the Ensenso test set. However, it still exhibits significant pose errors on certain objects. In contrast, our method maintains higher robustness and consistently delivers accurate pose estimations.

\newpage
\clearpage

\end{document}